\documentclass{article}

\usepackage[numbers, compress]{natbib}
\usepackage[margin=1.1in]{geometry}
\usepackage{authblk}

\usepackage[utf8]{inputenc} %
\usepackage[T1]{fontenc}    %
\usepackage{hyperref}       %
\usepackage{url}            %
\usepackage{booktabs}       %
\usepackage{amsfonts}       %
\usepackage{nicefrac}       %
\usepackage{microtype}      %

\usepackage[rgb]{xcolor}
\usepackage{bbm}
\usepackage{amsfonts}
\usepackage{algorithm}
\usepackage{algorithmic}
\usepackage{etoolbox}
\usepackage{dsfont}
\usepackage{amssymb}

\usepackage[cmex10]{amsmath}
\usepackage{amssymb}
\usepackage{graphicx}
\usepackage{wrapfig}
\usepackage[normalem]{ulem}

\usepackage{shortcut_conf}
\usepackage{etoolbox}

\usepackage{ stmaryrd }

\usepackage{subfigure}
\usepackage{caption}
\usepackage{multibib}
\newcites{latex}{References}

\usepackage[textwidth=2.4cm, textsize=scriptsize]{todonotes}

\newcommand{\mjtodo}[1]{\todo[color=orange!40]{{\bf Mainak:} #1}}
\newcommand\mainakst{\bgroup\markoverwith{\textcolor{red}{\rule[0.5ex]{2pt}{0.4pt}}}\ULon}

\newcommand{\tom}[1]{\textcolor{red}{{\bf Tom:} #1}}

\newtoggle{final}
\toggletrue{final}

\iftoggle{final}{
\renewcommand{\mjtodo}[1]{}
\renewcommand{\tom}[1]{}
}{}

\title{Learning the Morphology of Brain Signals Using Alpha-Stable Convolutional Sparse Coding}

\begin{document}

\author[1]{Mainak Jas\thanks{mainak.jas@telecom-paristech.fr}}
\author[1]{Tom Dupr\'e La Tour}
\author[1]{Umut \c Sim\c sekli}
\author[2]{Alexandre Gramfort}
\affil[1]{LTCI, T\'{e}l\'{e}com ParisTech, Universit\'e Paris-Saclay, Paris, France}
\affil[2]{INRIA, Universit\'e Paris Saclay, Saclay, France}
\renewcommand\Authands{ and }
\date{\vspace{-5ex}}
\maketitle

\begin{abstract}

Neural time-series data contain a wide variety of prototypical signal waveforms (atoms) that are of significant importance in clinical and cognitive research. One of the goals for analyzing such data is hence to extract such `shift-invariant' atoms. Even though some success has been reported with existing algorithms, they are limited in applicability due to their heuristic nature. Moreover, they are often vulnerable to artifacts and impulsive noise, which are typically present in raw neural recordings.
In this study, we address these issues and propose a novel probabilistic convolutional sparse coding (CSC) model for learning shift-invariant atoms from raw neural signals containing potentially severe artifacts. In the core of our model, which we call $\alpha$CSC, lies a family of heavy-tailed distributions called $\alpha$-stable distributions. We develop a novel, computationally efficient Monte Carlo expectation-maximization algorithm for inference. The maximization step boils down to a weighted CSC problem, for which we  develop a computationally efficient optimization algorithm.
Our results show that the proposed algorithm achieves state-of-the-art convergence speeds. Besides, $\alpha$CSC is significantly more robust to artifacts when compared to three competing algorithms: it can extract spike bursts, oscillations, and even reveal more subtle phenomena such as cross-frequency coupling when applied to noisy neural time series.

\end{abstract}

\section{Introduction}
Neural time series data, either non-invasive such as electroencephalograhy (EEG) 
or invasive such as electrocorticography (ECoG) and local field potentials (LFP), are fundamental to modern experimental neuroscience. Such recordings contain a wide variety of `prototypical signals' that range from beta rhythms (12--30 Hz) in motor imagery tasks and alpha oscillations (8--12 Hz) involved in attention mechanisms, to spindles in sleep studies, 
and the classical P300 event related potential, a biomarker for surprise. 
These prototypical waveforms are considered critical in clinical and cognitive research~\cite{cole2017brain}, thereby motivating the development of computational tools for learning such signals from data.

Despite the underlying complexity in the morphology of neural signals, the majority of the computational tools in the community are based on representing the signals with rather simple, predefined bases, such as the Fourier or wavelet bases~\cite{cohen2014analyzing}.
While such bases lead to computationally efficient algorithms, they often fall short at capturing the precise morphology of signal waveforms, as demonstrated by a number of recent studies~\cite{jones2016brain,mazaheri2008asymmetric}. An example of such a failure is the disambiguation of the alpha rhythm from the mu rhythm~\cite{hari2017meg}, both of which have a component around $10$\,Hz but with different morphologies that cannot be captured by Fourier- or wavelet-based representations.

Recently, there have been several attempts for extracting more realistic and precise morphologies directly from unfiltered electrophysiology signals, via dictionary learning approaches \cite{jost2006motif,brockmeier2016learning,hitziger2017adaptive,gips2017discovering}.
These methods all aim to extract certain \emph{shift-invariant} prototypical waveforms (called `atoms' in this context) to better capture the temporal structure of the signals.
As opposed to using generic bases that have predefined shapes, such as the Fourier or the wavelet bases, these atoms provide a more meaningful representation of the data and are not restricted to narrow frequency bands.

In this line of research, \citet{jost2006motif} proposed the MoTIF algorithm, which uses an iterative strategy based on generalized eigenvalue decompositions, where the atoms are assumed to be orthogonal to each other and learnt one by one in a greedy way.
More recently, the `sliding window matching' (SWM) algorithm \cite{gips2017discovering} was proposed for learning time-varying atoms by using a correlation-based approach that aims to identify the recurring patterns. Even though some success has been reported with these algorithms,
they have several limitations: SWM uses a slow stochastic search inspired by simulated annealing and MoTIF poorly handles correlated atoms, simultaneously activated, or having varying amplitudes; some cases which often occur in practical applications.

A natural way to cast the problem of learning a dictionary of shift-invariant atoms into an optimization problem is a convolutional sparse coding (CSC) approach~\cite{Grosse-etal:2007}. 
This approach has gained popularity in computer vision~\cite{heide2015fast,wohlberg2016efficient,zeiler2010deconvolutional, vsorel2016fast,kavukcuoglu2010learning}%
, biomedical imaging~\cite{pachitariu2013extracting} and audio signal processing~\cite{Grosse-etal:2007,mailhe2008shift},
due to its ability to obtain compact representations of the signals and to incorporate the temporal structure of the signals via convolution.
In the neuroscience context, \citet{barthelemy2013multivariate} used an extension of the K-SVD algorithm using convolutions on EEG data. 
In a similar spirit, \citet{brockmeier2016learning} used the matching pursuit algorithm combined with a rather heuristic  
dictionary update, which is similar to the MoTIF algorithm. 
In a very recent study, \citet{hitziger2017adaptive} proposed the AWL algorithm, which presents a mathematically more principled CSC approach for modeling neural signals. Yet, as opposed to classical CSC approaches, the AWL algorithm imposes additional combinatorial  constraints, which limit its scope to certain data that contain spike-like atoms. Also, since these constraints increase the complexity of the optimization problem, the authors had to resort to dataset-specific initializations and many heuristics in their inference procedure.

While the current state-of-the-art CSC methods have a strong potential for modeling neural signals, they might also be limited as they consider an $\ell_2$ reconstruction error, which corresponds to assuming an additive Gaussian noise distribution. While this assumption could be reasonable for several signal processing tasks, it turns out to be very restrictive for neural signals, which often contain heavy noise bursts and have low signal-to-noise ratio.

In this study, we aim to address the aforementioned concerns and propose a novel probabilistic CSC model called $\alpha$CSC, which is better-suited for neural signals. $\alpha$CSC is based on a family of \emph{heavy-tailed} distributions called $\alpha$-stable distributions \cite{samorodnitsky1994stable} whose rich structure covers a broad range of noise distributions. The heavy-tailed nature of the $\alpha$-stable distributions renders our model robust to impulsive observations. We develop a Monte Carlo expectation maximization (MCEM) algorithm for inference, with a weighted CSC model for the maximization step. We propose efficient optimization strategies that are specifically designed for neural time series.
We illustrate the benefits of the proposed approach on both synthetic and real datasets.

\section{Preliminaries}

\textbf{Notation:} For a vector $v \in \bbR^n$ we denote the $\ell_p$ norm by $\|v\|_p = \left(\sum_i |v_i|^p \right)^{1/p}$. The convolution of two vectors $v_1 \in \bbR^N$ and $v_2 \in \bbR^M$ is denoted by $v_1 \ast v_2 \in \bbR^{N + M - 1}$. We denote by $x$ the observed signals, $d$ the temporal atoms, and $z$ the sparse vector of \emph{activations}. The symbols ${\cal U}$, ${\cal E}$, ${\cal N}$, ${\cal S}$ denote the univariate uniform, exponential, Gaussian, and $\alpha$-stable distributions, respectively.

\textbf{Convolutional sparse coding:} 
The CSC problem formulation adopted in this work follows the Shift Invariant Sparse Coding (SISC) model from~\cite{Grosse-etal:2007}. It is defined as follows:
\begin{align}
 \min_{d, z} \sum_{n=1}^{N} \Big( \frac{1}{2}\|x_{n} - \sum_{k=1}^{K}d^{k} * z_{n}^{k}\|_{2}^{2} + \lambda \sum_{k=1}^K \|z_{n}^{k}\|_1 \Big), \hspace{9pt}
 \text{s.t. } \>\> \|d^{k}\|_2^2 \leq 1 \text{  and } z_n^k \geq 0, \forall n, k
\label{eq:problem_definition} \enspace ,
\end{align}
where $x_{n} \in \bbR^{T}$ denotes one of the $N$ observed segments of signals, also referred to as a \emph{trials} in this paper. We denote by $T$ as the length of a trial, and $K$ the number of atoms. The aim in this model is to approximate the signals $x_n$ by the convolution of certain \emph{atoms} and their respective \emph{activations}, which are sparse. Here, $d^{k} \in \bbR^{L}$ denotes the $k$th atom of the \emph{dictionary} $d \equiv \{d^k\}_{k}$, and $z_{n}^{k} \in \bbR_+^{T-L+1}$ denotes the activation of the $k$th atom in the $n$th trial. We denote by $z \equiv \{z_n^k\}_{n,k}$.

The objective function \eqref{eq:problem_definition} has two terms, an $\ell_2$ data fitting term that corresponds to assuming an additive Gaussian noise model, and a regularization term that promotes sparsity with an $\ell_1$ norm.
The regularization parameter is called $\lambda > 0$. Two constraints are also imposed. First, we ensure that $d^{k}$ lies within the unit sphere, which prevents the scale ambiguity between $d$ and $z$. Second, a positivity constraint on $z$ is imposed to be able to obtain physically meaningful activations and to avoid sign ambiguities between $d$ and $z$. This positivity constraint is not present in the original SISC model~\cite{Grosse-etal:2007}.

\begin{figure}[t]
    \centering
    \subfigure[]{
    \includegraphics[width=0.31\linewidth]{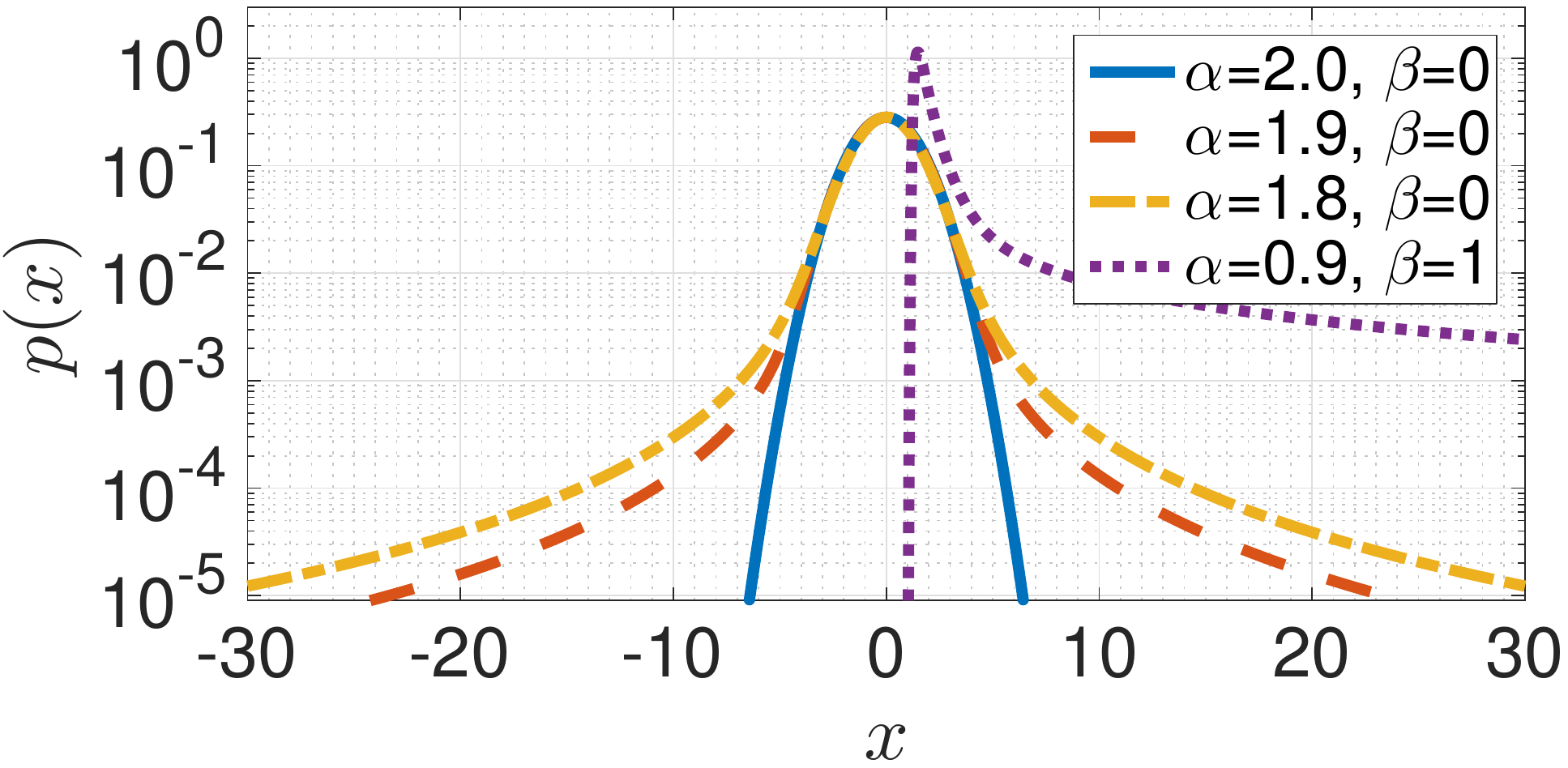}
    \label{fig:stable_pdf}
    } \hfill
    \subfigure[]{
    \includegraphics[width=0.31\linewidth]{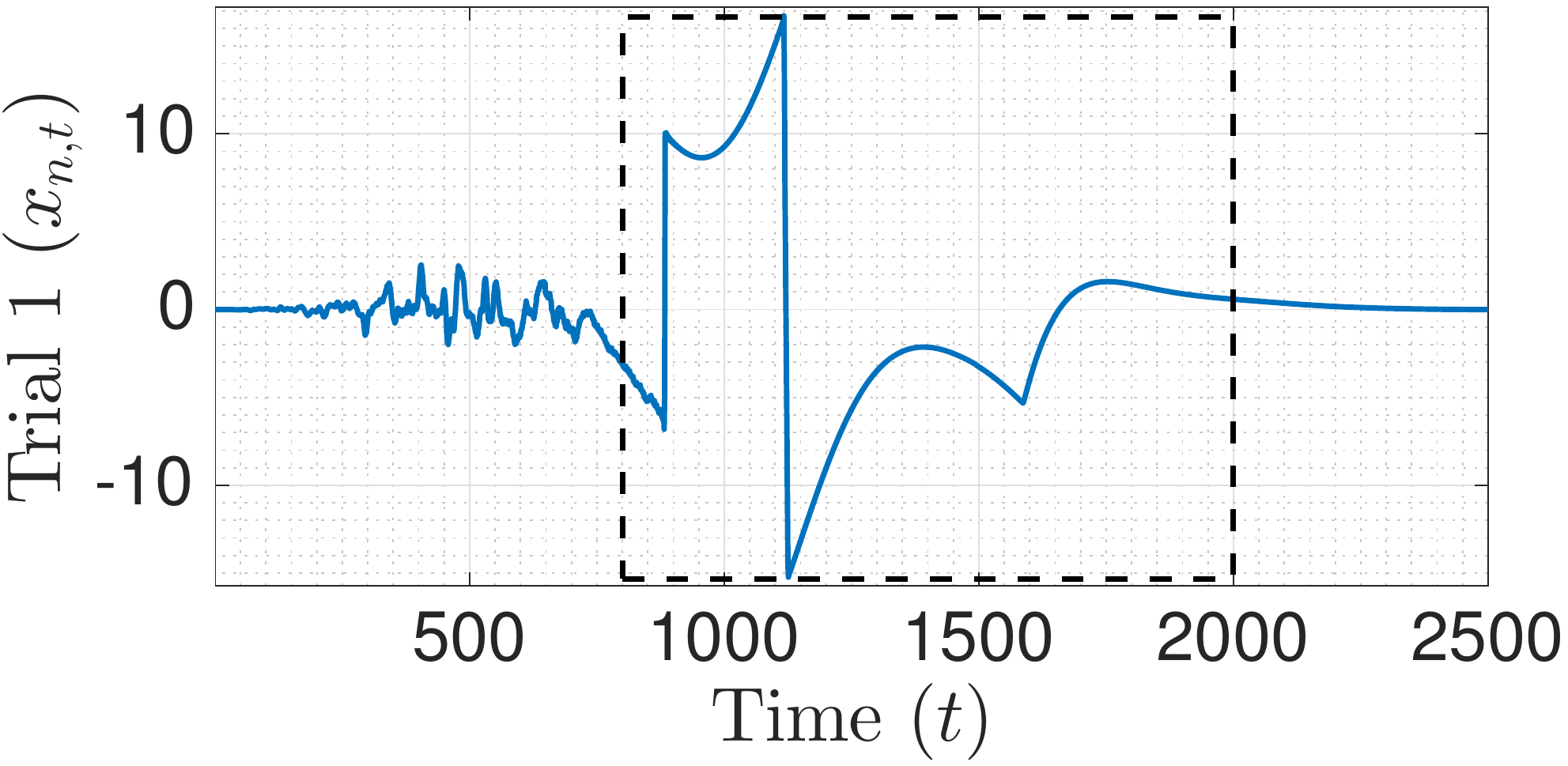}
    \includegraphics[width=0.31\linewidth]{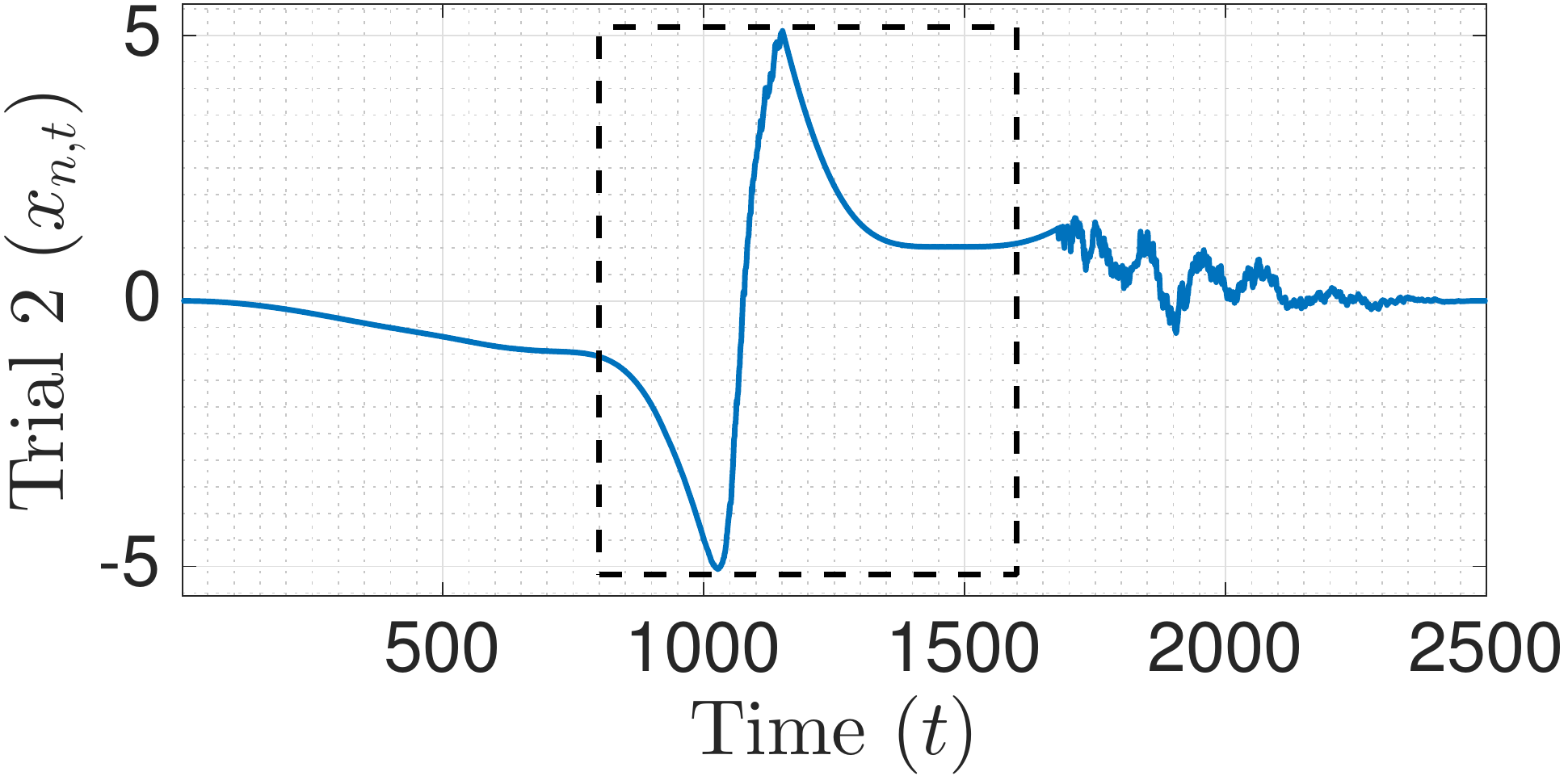}
    \label{fig:artifacts}
    }
    \vspace{-10pt}
    \caption{(a) PDFs of $\alpha$-stable distributions. (b) Illustration of two trials from the striatal LFP data, which contain severe artifacts. The artifacts are illustrated with dashed rectangles.}
    \label{fig:pdf_lfp}
\end{figure}

\textbf{$\alpha$-Stable distributions:} 
The $\alpha$-stable distributions have become increasingly popular in modeling signals that might incur large variations \cite{kuruoglu1999signal, mandelbrot2013fractals, wang2016delving} and have a particular importance in statistics since they appear as the limiting distributions in the generalized central limit theorem \cite{samorodnitsky1994stable}. They are characterized by four parameters: $\alpha$, $\beta$, $\sigma$, and $\mu$:
(i) $\alpha \in (0,2]$ is the \emph{characteristic exponent} and determines the tail thickness of the distribution: the distribution will be heavier-tailed as $\alpha$ gets smaller. 
(ii) $\beta \in [-1 ,1]$ is the \emph{skewness} parameter.
If $\beta = 0$, the distribution is symmetric.
(iii) $\sigma \in (0,\infty)$ is the \emph{scale} parameter and measures the spread of the random variable around its mode. Finally, (iv) $\mu \in (-\infty, \infty)$ is the location parameter. 

The probability density function of an $\alpha$-stable distribution cannot be written in closed-form except for certain special cases; however, the characteristic function  can be written as follows:
\begin{align*}
x \sim {\cal S}(\alpha,\beta,\sigma,\mu) \iff \mathds{E}[\exp( i \omega x)]  = \exp(-|\sigma \omega|^\alpha \left[1+ i \sign(\omega)\beta \psi_\alpha(\omega)  \right] + i \mu \omega ) \enspace ,
\end{align*}
where $\psi_\alpha(\omega) = \log |\omega| $ for $\alpha =1$, $\psi_\alpha(\omega) = \tan(\pi \alpha/2)$ for $\alpha \neq 1$, and $i = \sqrt{-1}$. 
As an important special case of the $\alpha$-stable distributions, we obtain the Gaussian distribution when $\alpha = 2$ and $\beta =0$, \textit{i.e.}\ ${\cal S}(2,0,\sigma,\mu) = {\cal N}(\mu,2 \sigma^2)$. 
In Fig.~\ref{fig:stable_pdf}, we illustrate the (approximately computed) probability density functions (PDF) of the $\alpha$-stable distribution for different values of $\alpha$ and $\beta$. The distribution becomes heavier-tailed as we decrease $\alpha$, whereas the tails vanish quickly when $\alpha=2$.

The moments of the $\alpha$-stable distributions can only be defined up to the order $\alpha$, i.e. $\mathds{E}[|x|^p] < \infty $ if and only if $p <\alpha$, which implies the distribution has infinite variance when $\alpha<2$. Furthermore, despite the fact that the PDFs of $\alpha$-stable distributions do not admit an analytical form, it is straightforward to draw random samples from $\alpha$-stable distributions~\cite{chambers1976method}.

\section{Alpha-Stable Convolutional Sparse Coding}

\subsection{The Model}

From a probabilistic perspective, the CSC problem can be also formulated as a maximum a-posteriori (MAP) estimation problem on the following probabilistic generative model:
\begin{align}
z_{n,t}^k \sim {\cal E}(\lambda),
\quad x_{n,t} | z, d \sim {\cal N}( \hat{x}_{n,t},1 ),
\quad \text{ where,}
\quad \hat{x}_n \triangleq \sum_{k=1}^{K}d^{k} * z_{n}^{k} \enspace .
\label{eqn:csc_prob}
\end{align}
Here, $z_{n,t}^k$ denotes the $t$th element of $z_{n}^k$. We use the same notations for $x_{n,t}$ and $\hat{x}_{n,t}$. It is easy to verify that the MAP estimate for this probabilistic model, \textit{i.e.}\ $\max_{d,z} \log p(d,z|x)$, is identical to the original optimization problem defined in~\eqref{eq:problem_definition}.

It has been long known that, due to their light-tailed nature, Gaussian models often fail at handling noisy high amplitude observations or outliers~\cite{Huber81a}. As a result, the `vanilla' CSC model turns out to be highly sensitive to outliers and impulsive noise that frequently occur in electrophysiological recordings, as illustrated in Fig.~\ref{fig:artifacts}. Possible origins of such artifacts are movement, muscle contractions, ocular blinks or electrode contact losses.

In this study, we aim at developing a probabilistic CSC model that would be capable of modeling challenging electrophysiological signals. We propose an extension of the original CSC model defined in~\eqref{eqn:csc_prob} by replacing the light-tailed Gaussian likelihood with heavy-tailed $\alpha$-stable distributions. We define the proposed probabilistic model ($\alpha$CSC) as follows:
\begin{align}
z_{n,t}^k \sim {\cal E}( \lambda),  \quad
x_{n,t} | z, d \sim {\cal S} (\alpha, 0, 1/\sqrt{2}, \hat{x}_{n,t} ) \enspace , \label{eqn:acsc_org}
\end{align}
where ${\cal S}$ denotes the $\alpha$-stable distribution. 
While still being able to capture the temporal structure of the observed signals via convolution, the proposed model has a richer structure and would allow large variations and outliers, thanks to the heavy-tailed $\alpha$-stable distributions. Note that the vanilla CSC defined in \eqref{eqn:csc_prob} appears as a special case of $\alpha$CSC, as the $\alpha$-stable distribution coincides with the Gaussian distribution when $\alpha=2$.

\subsection{Maximum A-Posteriori Inference}
Given the observed signals $x$, we are interested in the MAP estimates, defined as follows:
\begin{align}
(d^\star,z^\star) = \argmax_{d,z}  \sum_{n,t} \Bigl( \log p(x_{n,t}|d,z) + \sum_k \log p(z_{n,t}^k)  \Bigr).
\end{align}
As opposed to the Gaussian case, unfortunately, this optimization problem is not amenable to classical optimization tools, since the PDF of the $\alpha$-stable distributions does not admit an analytical expression.  
As a remedy, we use the product property of the symmetric $\alpha$-stable densities \cite{samorodnitsky1994stable,godsill1999bayesian} and re-express the $\alpha$CSC model as conditionally Gaussian. It leads to:
\begin{align}
z_{n,t}^k \sim {\cal E}( \lambda),  \quad 
\phi_{n,t} \sim {\cal S}\Bigl(\frac{\alpha}{2},1, 2 (\cos \frac{\pi \alpha}{4})^{2/\alpha} ,0 \Bigr), \quad
x_{n,t} | z, d, \phi \sim {\cal N}\Bigl(\hat{x}_{n,t},\frac{1}{2}\phi_{n,t} \Bigr) \enspace ,
\label{eqn:sas_condgauss}
\end{align}
where $\phi$ is called the \emph{impulse} variable that is drawn from a \emph{positive} $\alpha$-stable distribution (i.e.\ $\beta =1$), whose PDF is illustrated in Fig.~\ref{fig:stable_pdf}. It can be easily shown that both formulations of the $\alpha$CSC model are identical by marginalizing the joint distribution $p(x,d,z,\phi)$ over $\phi$. 

The impulsive structure of the $\alpha$CSC model becomes more prominent in this formulation: the variances of the Gaussian observations are modulated by stable random variables with infinite variance, where the impulsiveness depends on the value of $\alpha$. 
It is also worth noting that when $\alpha = 2$, $\phi_{n,t}$ becomes deterministic and we can again verify that $\alpha$CSC coincides with the vanilla CSC.

The conditionally Gaussian structure of the augmented model has a crucial practical implication: if the impulse variable $\phi$ were to be known, then the MAP estimation problem over $d$ and $z$ in this model would turn into a `weighted' CSC problem, which is a much easier task compared to the original problem. In order to be able to exploit this property, we propose an expectation-maximization (EM) algorithm, which iteratively maximizes a lower bound of the log-posterior $\log p(d,z|x)$, and algorithmically boils down to computing the following steps in an iterative manner:
\begin{align}
&\text{E-Step:} \hspace{20pt} {\cal B}^{(i)}(d,z) = \mathds{E}\left[\log p(x,\phi,z|d)\right]_{p(\phi|x,z^{(i)},d^{(i)})}, \\
&\text{M-Step:} \hspace{20pt} (d^{(i+1)}, z^{(i+1)}) = \argmax\nolimits_{d,z} {\cal B}^{(i)}(d,z). \label{eq:mstep}
\end{align}
where $\mathds{E}[f(x)]_{q(x)}$ denotes the expectation of a function $f$ under the distribution $q$, $i$ denotes the iterations, and ${\cal B}^{(i)}$ is a lower bound to $\log p(d,z|x)$ and it is tight at the current iterates $z^{(i)}$, $d^{(i)}$.

\textbf{The E-Step:} 
In the first step of our algorithm, we need to compute the EM lower bound ${\cal B}$ that has the following form:
\begin{align}
{\cal B}^{(i)}(d,z) =^+ - \sum_{n=1}^N \Big( \|\sqrt{w_{n}^{(i)}} \odot (x_{n} - \sum_{k=1}^{K}d^{k} * z_{n}^{k})\|_{2}^{2} + \lambda \sum_{k=1}^K{\|z_{n}^{k}\|_1}\Big),
\end{align}
where $=^+$ denotes equality up to additive constants, $\odot$ denotes the Hadamard (element-wise) product, and the square-root operator is also defined element-wise. Here, $w_{n}^{(i)} \in \bbR^T_+$ are the \emph{weights} that are defined as follows: $w_{n,t}^{(i)} \triangleq \mathds{E}\left[1/{\phi_{n,t}}\right]_{p(\phi|x,z^{(i)},d^{(i)})}$. As the variables $\phi_{n,t}$ are expected to be large when $\hat{x}_{n,t}$ cannot explain the observation $x_{n,t}$ -- typically due to a corruption or a high noise -- the weights will accordingly suppress the importance of the particular point $x_{n,t}$. Therefore, the overall approach will be more robust to corrupted data than the Gaussian models where all weights would be deterministic and equal to $0.5$. 

\begin{wrapfigure}{R}{0.52\textwidth}
\vspace{-10pt}
    \begin{minipage}{0.52\textwidth}
    \begin{algorithm}[H]
      \begin{algorithmic}[1] %
      \REQUIRE Regularization: $\lambda \in \real_+$, Num. atoms: $K$, Atom length: $L$, Num. iterations: $I$ , $J$, $M$
        \FOR{$i=1$ to $I$}
          \STATE \textit{\color{blue} /* E-step: */} %
          \FOR{$j=1$ to $J$}
          \STATE Draw $\phi_{n,t}^{(i,j)}$ via MCMC \eqref{eqn:mcmc_acc}
          \ENDFOR
          \STATE $w_{n,t}^{(i)} \approx (1/J) \sum\nolimits_{j=1}^{J} 1/{\phi_{n,t}^{(i,j)}}$
          \STATE \textit{\color{blue} /* M-step: */} %
              \FOR{$m=1$ to $M$}
                  \STATE $z^{(i)}$ = L-BFGS-B on \eqref{eq:problem_definition_z}
                  \STATE $d^{(i)}$ = L-BFGS-B on the dual of \eqref{eq:problem_definition_d}
              \ENDFOR
        \ENDFOR
        \RETURN $w^{(I)}$, $d^{(I)}$, $z^{(I)}$
        \end{algorithmic}
        \caption{$\alpha$-stable Convolutional Sparse Coding}
        \label{alg:alpha_csc}
    \end{algorithm}
\end{minipage}
\vspace{-20pt}
\end{wrapfigure}
Unfortunately, the weights $w^{(i)}$ cannot be computed analytically, therefore we need to resort to
approximate methods. In this study, we develop a Markov chain Monte Carlo (MCMC) method to approximately compute the weights, where we approximate the intractable expectations with a finite sample average, given as follows: $w_{n,t}^{(i)} \approx (1/{J}) \sum_{j=1}^{J} 1/{\phi_{n,t}^{(i,j)}}$, where $\phi_{n,t}^{(i,j)}$ are some samples that are ideally drawn from the posterior distribution $p(\phi|x,z^{(i)},d^{(i)})$. Unfortunately, directly drawing samples from the posterior distribution of $\phi$ is not tractable either, and therefore, we develop a \emph{Metropolis-Hastings} algorithm \cite{chib1995understanding}, that asymptotically generates samples from the \emph{target} distribution $p(\phi|\cdot)$ in two steps. In the $j$-th iteration of this algorithm, we first draw a random sample for each $n$ and $t$ from the prior distribution (cf.\ \eqref{eqn:sas_condgauss}), \textit{i.e.}, $\phi_{n,t}'\sim p(\phi_{n,t})$. We then compute an acceptance probability for each $\phi_{n,t}'$ that is defined as follows:
\begin{align}
  \text{acc}(\phi_{n,t}^{(i,j)} \rightarrow \phi_{n,t}' ) \triangleq \min \Bigl\{1, {p(x_{n,t}|d^{(i)},z^{(i)},\phi'_{n,t})}/{p(x_{n,t}|d^{(i)},z^{(i)},\phi_{n,t}^{(i,j)})} \Bigr\} \label{eqn:mcmc_acc}
\end{align}
where $j$ denotes the iteration number of the MCMC algorithm. 
Finally, we draw a uniform random number $u_{n,t} \sim {\cal U}([0, 1])$ for each $n$ and $t$. If $u_{n,t} < \text{acc}(\phi_{n,t}^{(i)} \rightarrow \phi_{n,t}')$, we accept the sample and set $\phi_{n,t}^{(i+1)} = \phi_{n,t}'$; otherwise we reject the sample and set $\phi_{n,t}^{(i+1)} = \phi_{n,t}^{(i)}$. This procedure forms a Markov chain that leaves the target distribution $p(\phi|\cdot)$ invariant, where under mild ergodicity conditions, it can be shown that the finite-sample averages converge to their true values when $J$ goes to infinity \cite{Liu2008}. More detailed explanation of this procedure is given in the supplementary document.

\textbf{The M-Step:} 
Given the weights $w_n$ that are estimated during the E-step, the objective of the M-step~\eqref{eq:mstep} is to solve a weighted CSC problem, which is much easier when compared to our original problem. 
This objective function is not jointly convex in $d$ and $z$, yet it is convex if one fix
either $d$ or $z$.
Here, similarly to the vanilla CSC approaches~\cite{gips2017discovering,Grosse-etal:2007}, we develop a \emph{block coordinate descent} strategy, where we solve the problem in~\eqref{eq:mstep} for either $d$ or $z$, by keeping respectively $z$ and $d$ fixed.
We first focus on solving the problem for $z$ while keeping $d$ fixed, given as follows:
\begin{align}
& \min_{z} \sum_{n=1}^{N} \Big( \|\sqrt{w_{n}} \odot (x_{n} - \sum_{k=1}^{K}D^{k} \bar{z}_{n}^{k})\|_{2}^{2} + \lambda \sum_{k}{ \|{z}_{n}^{k} \|_1}\Big) \quad \text{ s.t.  } {z}_n^k \geq 0, \forall n,k\enspace .
\label{eq:problem_definition_z}
\end{align} 
Here, we expressed the convolution of $d^k$ and $z_n^k$ as the inner product of the zero-padded activations $\bar{z}_n^k \triangleq [(z_n^k)^\top, 0 \cdots 0]^\top \in \bbR^{T}_+$, with a Toeplitz matrix $D^k \in \bbR^{T \times T}$, that is constructed from $d^k$.
The matrices $D^k$ are never constructed in practice, and all operations are carried out using convolutions.
This problem can be solved by various constrained optimization algorithms. Here, we choose the quasi-Newton L-BFGS-B algorithm~\cite{byrd1995limited} with a box constraint: $0 \leq z_{n,t}^k \leq \infty$. This approach only requires the simple computation of the gradient of the objective function with respect to $z$ (\textit{cf.} supplementary material). Note that, since each trial is independent from each other, we can solve this problem for each $z_n$ in parallel. 

We then solve the problem for the atoms $d$ while keeping $z$ fixed. This optimization problem turns out to be a constrained weighted least-squares problem. In the non-weighted case, this problem can be solved either in the time domain or in the Fourier domain~\cite{Grosse-etal:2007,heide2015fast,wohlberg2016efficient}. The Fourier transform simplifies the convolutions that appear in least-squares problem, but it also induces several difficulties, such as that the atom $d_k$ have to be in a finite support $L$, an important issue ignored in the seminal work of~\cite{Grosse-etal:2007} and addressed with an ADMM solver  in\cite{heide2015fast,wohlberg2016efficient}.
In the weighted case, it is not clear how to solve this problem in the Fourier domain. We thus perform all the computations in the time domain.

Following the traditional filter identification approach~\cite{moulines1995subspace}, we need to embed the one-dimensional signals $z_n^k$ into a matrix of delayed signals $Z_n^{k} \in \bbR^{T \times L}$, where $ (Z_n^{k})_{i,j} = z_{n,i + j - L + 1}^k$ if $ L - 1 \le i+j < T$ and $0$ elsewhere. Equation~\eqref{eq:problem_definition} then becomes:
\begin{align}
& \min_{d} \sum_{n=1}^{N} \|\sqrt{w_n} \odot (x_{n} - \sum_{k=1}^{K}Z_{n}^{k}d^{k})\|_{2}^{2}, \quad \text{  s.t.  } \|d^k\|_2^2 \leq 1 \enspace.
\label{eq:problem_definition_d}
\end{align}
Due to the constraint, we must resort to an iterative approach. The options are to use (accelerated) projected gradient methods such as FISTA~\cite{beck2009fast} applied to~\eqref{eq:problem_definition_d}, or to solve a dual problem as done in~\cite{Grosse-etal:2007}. The dual is also a smooth constraint problem yet with a simpler positivity box constraint (\textit{cf.} supplementary material). The dual can therefore be optimized with L-BFGS-B. Using such a quasi-Newton solver turned out to be more efficient than any accelerated first order method in either the primal or the dual (\textit{cf.} benchmarks in supplementary material)

Our entire EM approach can be summarized in the Algorithm~\ref{alg:alpha_csc}.
Note that during the alternating minimization, thanks to convexity we can warm start the $d$ update and the $z$ update using the solution from the previous update. This significantly speeds up the convergence of the L-BFGS-B algorithm, particularly in the later iterations of the overall algorithm.

\begin{figure}[t]
    \centering
     \subfigure[$K=10$, $L=32$.]{
     \includegraphics[width=0.30\linewidth]{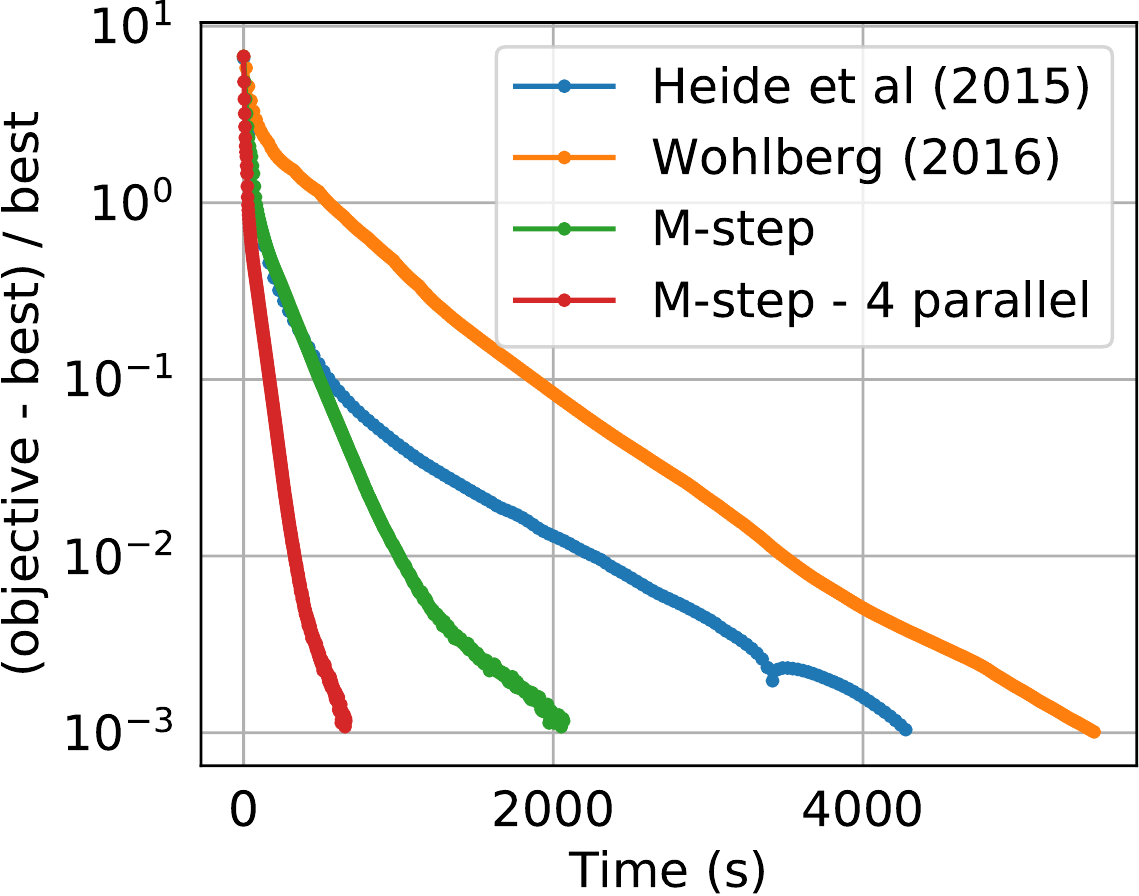}}
     \subfigure[Time to reach a relative precision of 0.01.]{
     \includegraphics[width=0.67\textwidth]{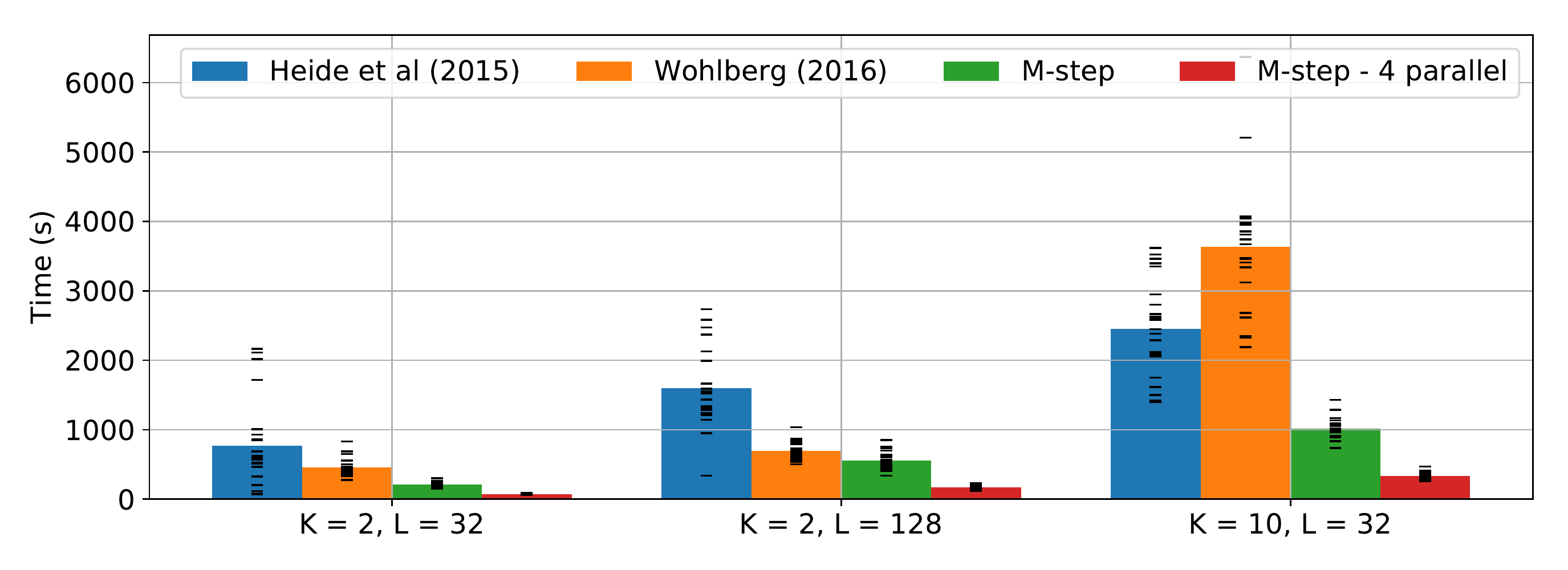}}
    \vspace{-10pt}
    \caption{Comparison of state-of-the-art methods with our approach. (a)~Convergence plot with the objective function relative to the obtained minimum, as a function of computational time. (b)~Time taken to reach a relative precision of $10^{-2}$, for different settings of $K$ and $L$.  }
    \label{fig:convergence}
\end{figure}

\section{Experiments}
\label{sec:experiments}
In order to evaluate our approach, we conduct several experiments on both synthetic and real data. 
First, we show that our proposed optimization scheme for the M-step provides significant improvements in terms
of convergence speed over the state-of-the-art CSC methods. Then, we provide empirical evidence that our algorithm is more robust to
artifacts and outliers than three competing CSC methods~\cite{jost2006motif,brockmeier2016learning,wohlberg2016efficient}. 
Finally, we consider LFP data, where we illustrate that our algorithm can reveal interesting properties in electrophysiological signals
without supervision, even in the presence of severe artifacts.

\textbf{Synthetic simulation setup:} 
In our synthetic data experiments, we simulate $N$ trials of length $T$ by first generating $K$ zero mean and unit norm atoms of length $L$. The  activation instants are integers drawn from a uniform distribution in $\llbracket0, T-L \rrbracket$. The amplitude of the activations are drawn from a uniform distribution in $[0, 1]$. Atoms are activated only once per trial and are allowed to overlap. The activations are then convolved with the generated atoms and summed up as in \eqref{eq:problem_definition}.

\textbf{M-step performance:} 
In our first set of synthetic experiments, we illustrate the benefits of our M-step optimization approach over state-of-the-art CSC solvers. 
We set $N=100$, $T=2000$ and $\lambda=1$, and use different values for $K$ and $L$. To be comparable, we set $\alpha=2$ and add Gaussian noise to the synthesized signals, where the standard deviation is set to $0.01$. In this setting, we  have $w_{n,t}=1/2$ for all $n$, $t$, which reduces the problem to a standard CSC setup. We monitor the convergence of ADMM-based methods by \citet{heide2015fast} and \citet{wohlberg2016efficient} against our M-step algorithm, using both a single-threaded and a parallel version for the $z$-update. 
As the problem is non-convex, even if two algorithms start from the same point, they are not guaranteed to reach the same local minimum. Hence, for a fair comparison, we use a multiple restart strategy with averaging across $24$ random seeds.

During our experiments we have observed that the ADMM-based methods do not guarantee the feasibility of the iterates. In other words, the norms of the estimated atoms might be greater than $1$ during the iterations. To keep the algorithms comparable, when computing the objective value, we project the atoms to the unit ball and scale the activations accordingly. To be strictly comparable, we also imposed a positivity constraint on these algorithms. This is easily done by modifying the soft-thresholding operator to be a rectified linear function. In the benchmarks, all algorithms use a single thread, except ``M-step - 4 parallel'' which uses 4 threads during the $z$ update.

In Fig.~\ref{fig:convergence}, we illustrate the convergence behaviors of the different methods.
Note that the y-axis is the precision relative to the objective value obtained upon convergence. In other words, each curve is relative to its own local minimum (see supplementary document for details).
In the right subplot, we show how long it takes for the algorithms to reach a relative precision of $0.01$ for different settings (\textit{cf.} supplementary material for more benchmarks). Our method consistently performs better and the difference is even more striking for more challenging setups. This speed improvement on the M-step is crucial for us as this step will be repeatedly executed. %
\newcommand{\tmpsize}{0.14}
\begin{figure}[t]
\begin{center}
\vspace{-10pt}
\subfigure[No corruption.]{
 \setlength{\tabcolsep}{0.5pt} 
    \begin{tabular}{c c}

        \includegraphics[width=\tmpsize\linewidth]{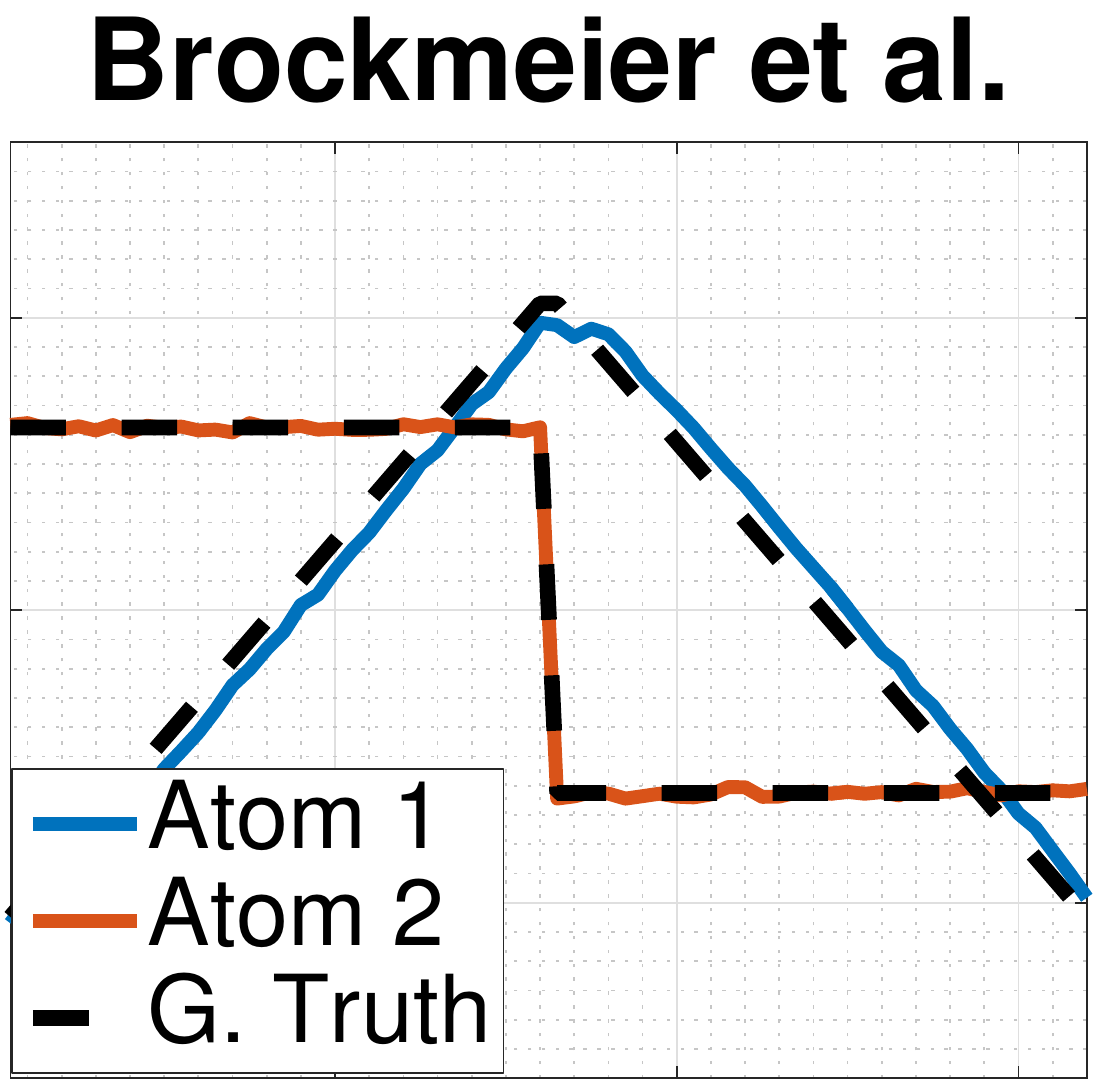} &
        \includegraphics[width=\tmpsize\linewidth]{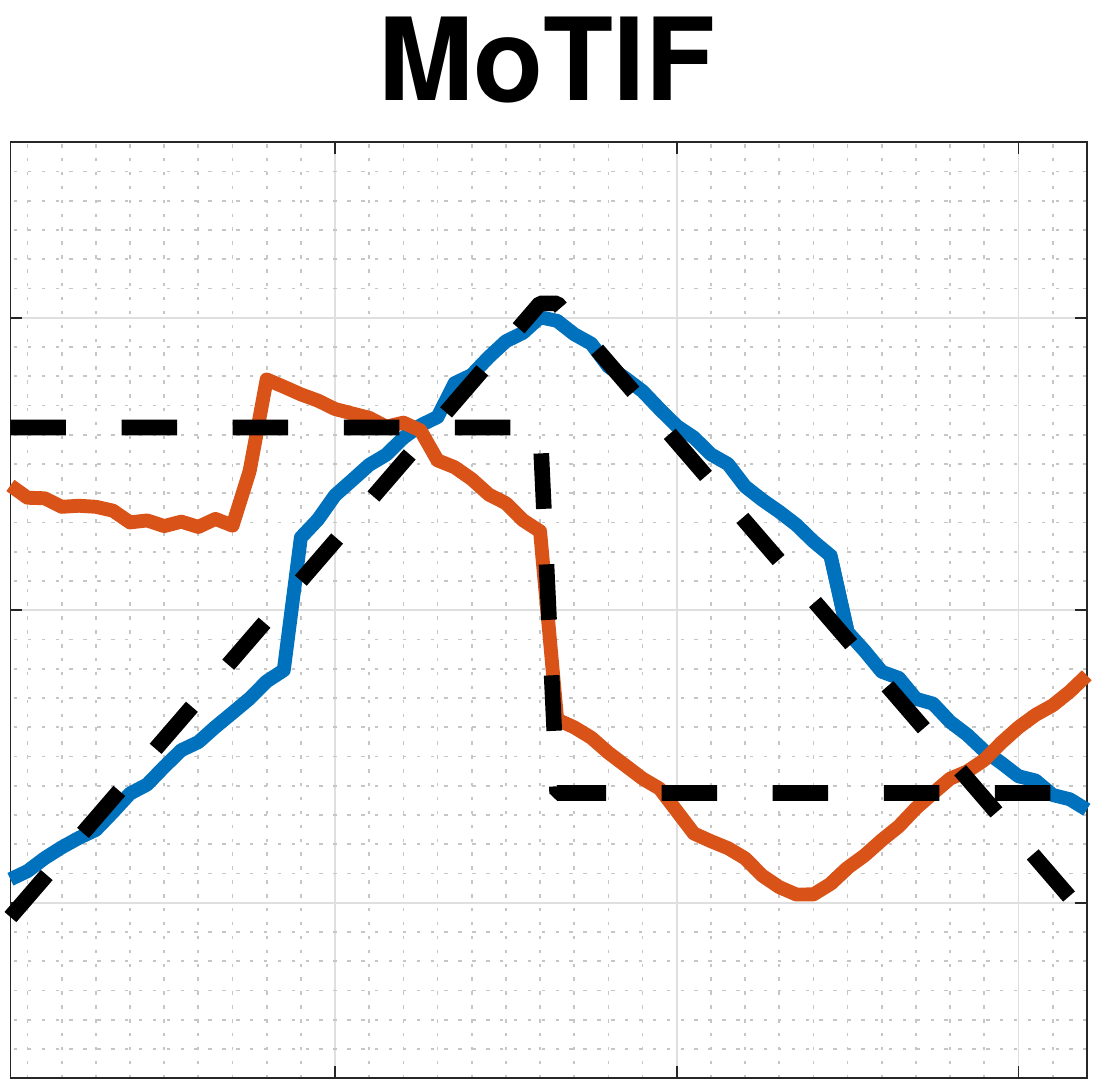}
        \\
        \includegraphics[width=\tmpsize\linewidth]{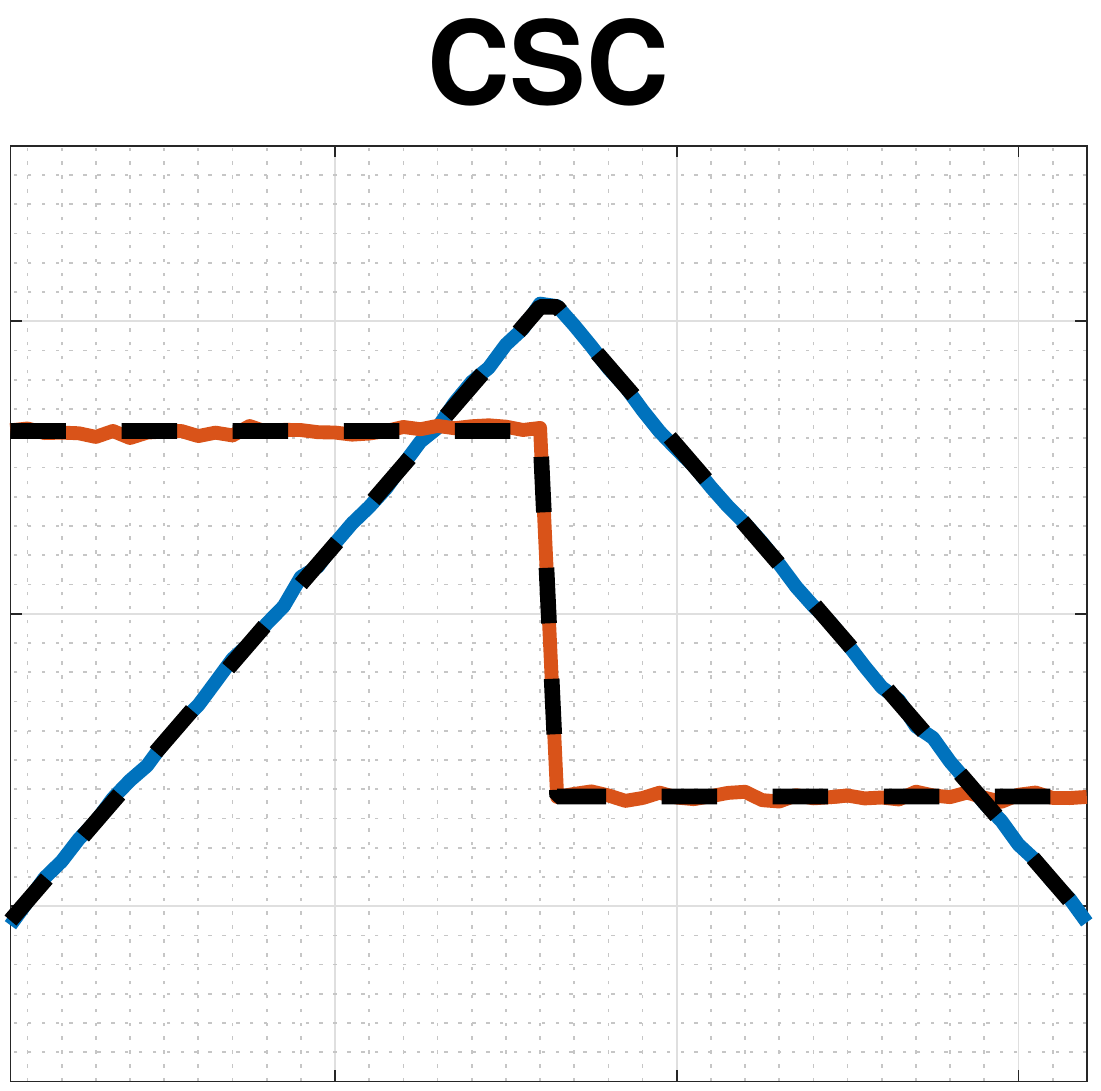} &
        \includegraphics[width=\tmpsize\linewidth]{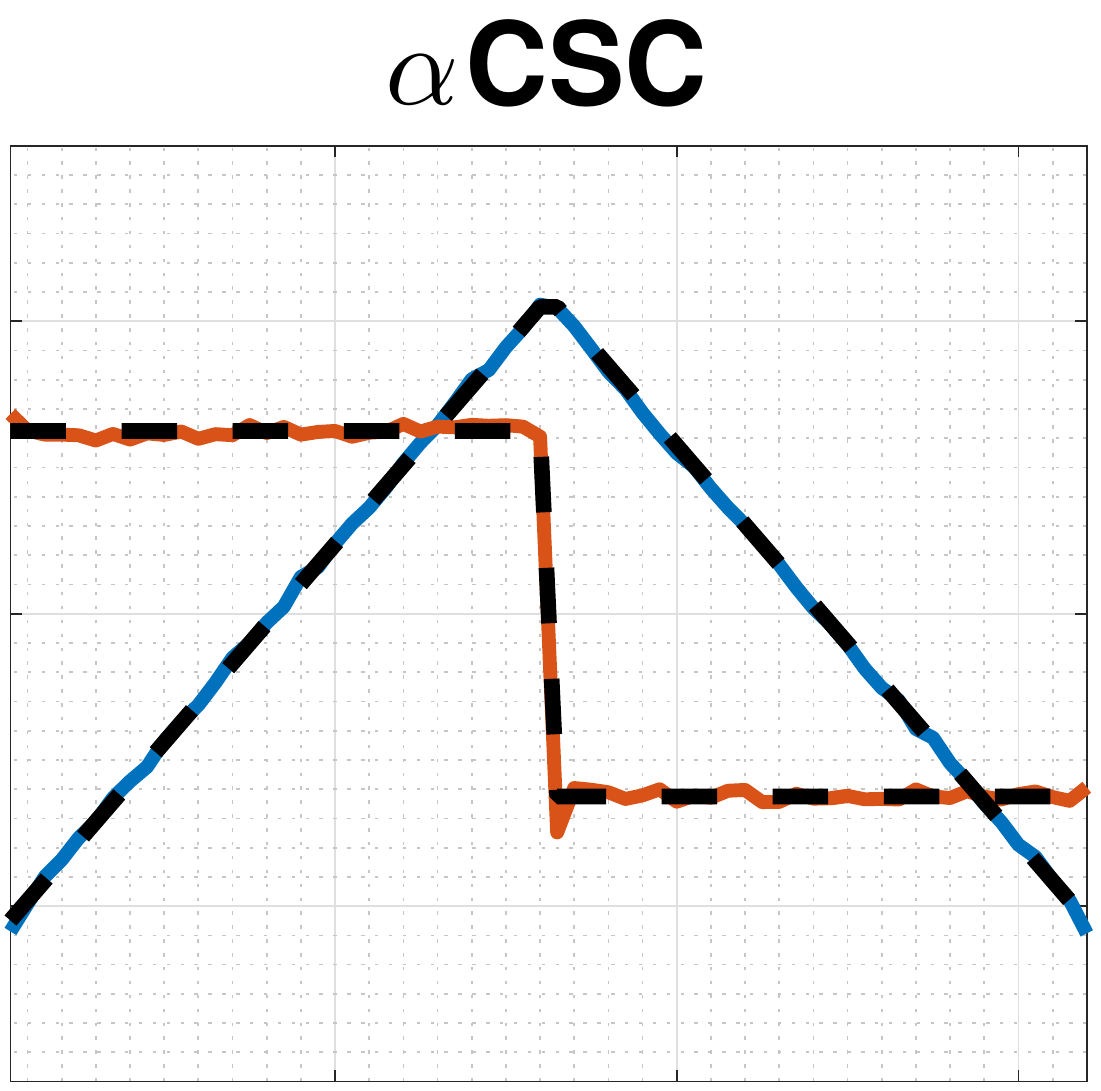}
        \end{tabular}
} \hfill
\subfigure[10\% corruption. ]{
\setlength{\tabcolsep}{0.5pt} 
\begin{tabular}{c c}
        \includegraphics[width=\tmpsize\linewidth]{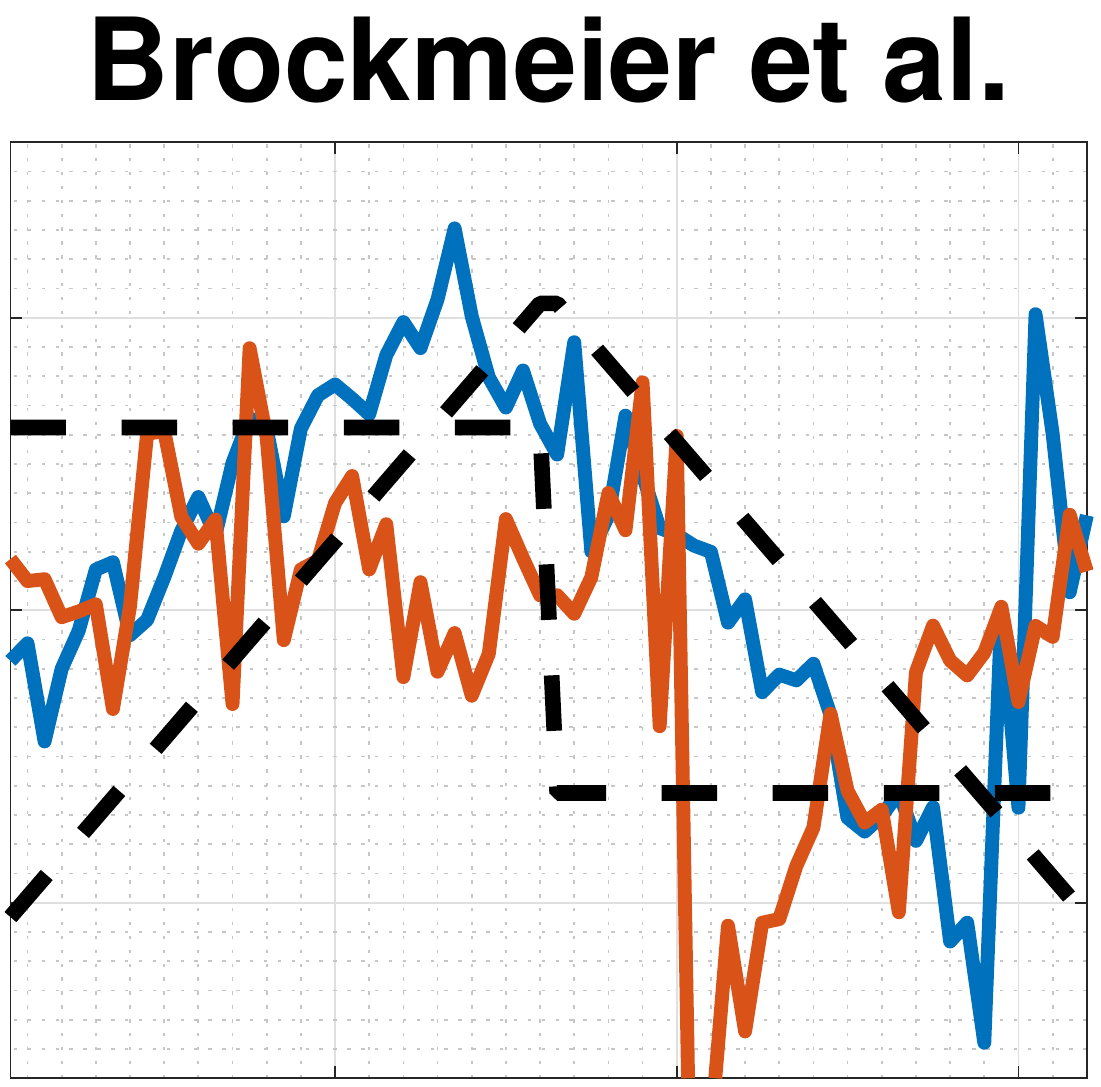} &
        \includegraphics[width=\tmpsize\linewidth]{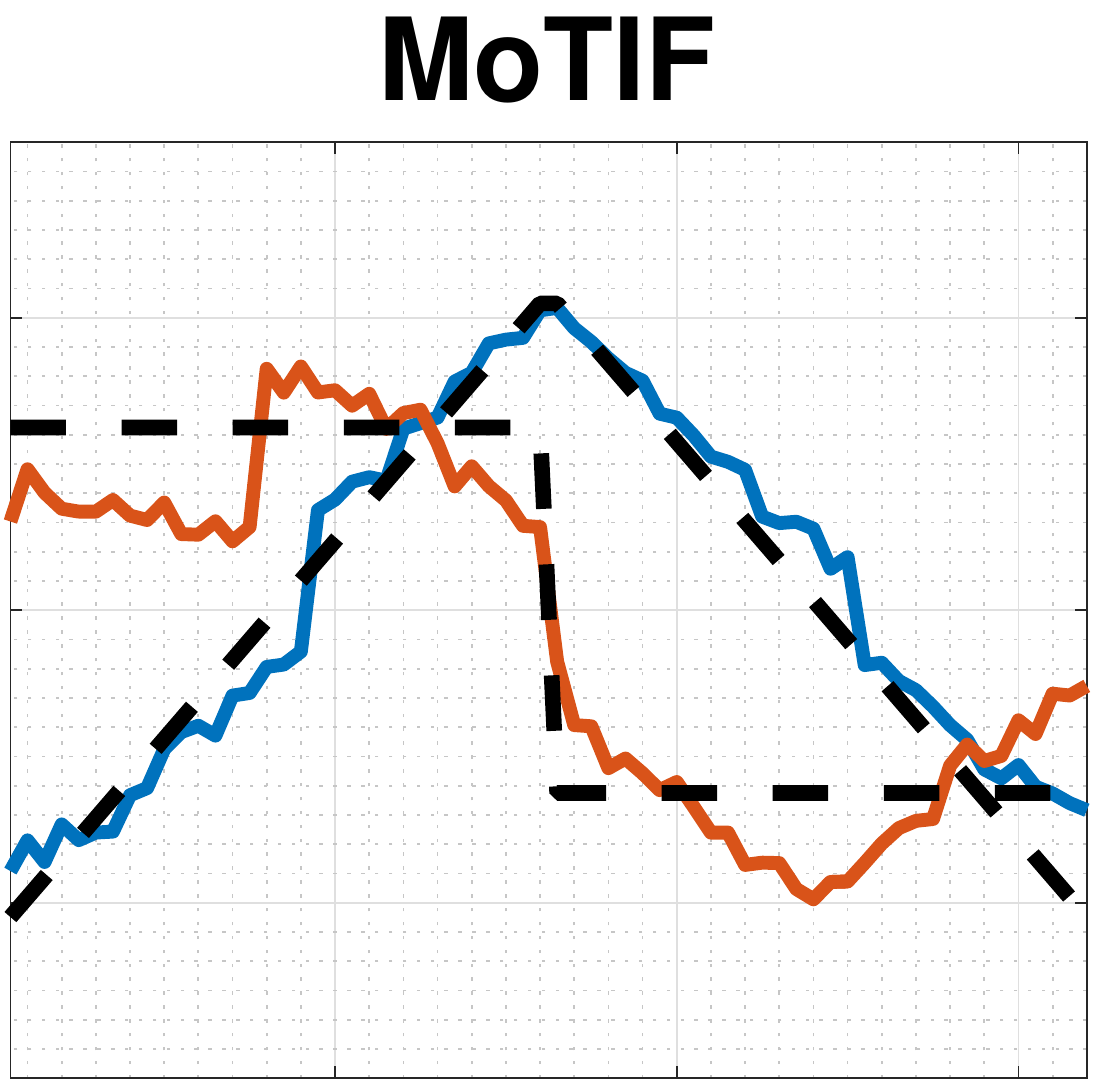}
\\
        \includegraphics[width=\tmpsize\linewidth]{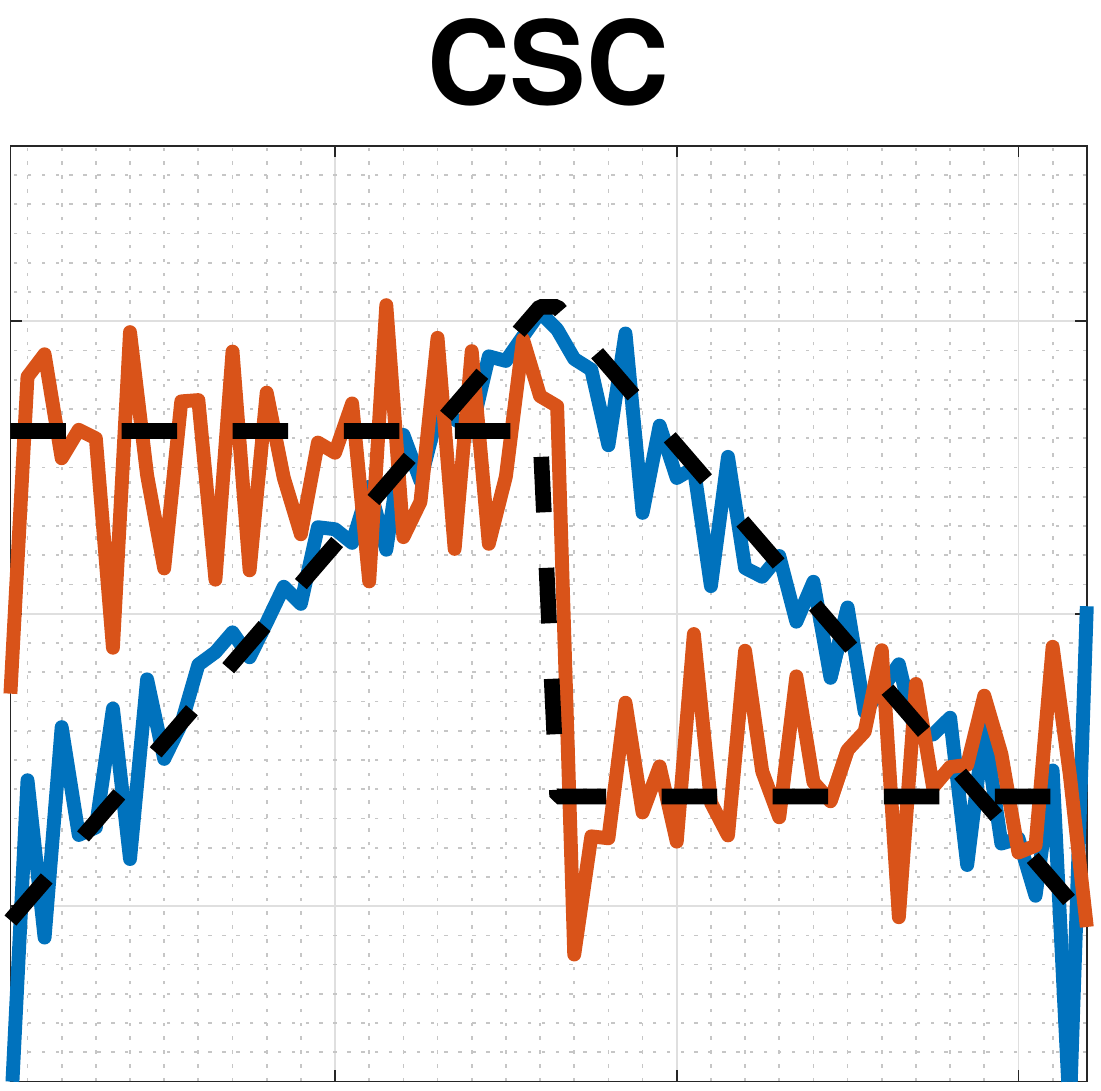} &
        \includegraphics[width=\tmpsize\linewidth]{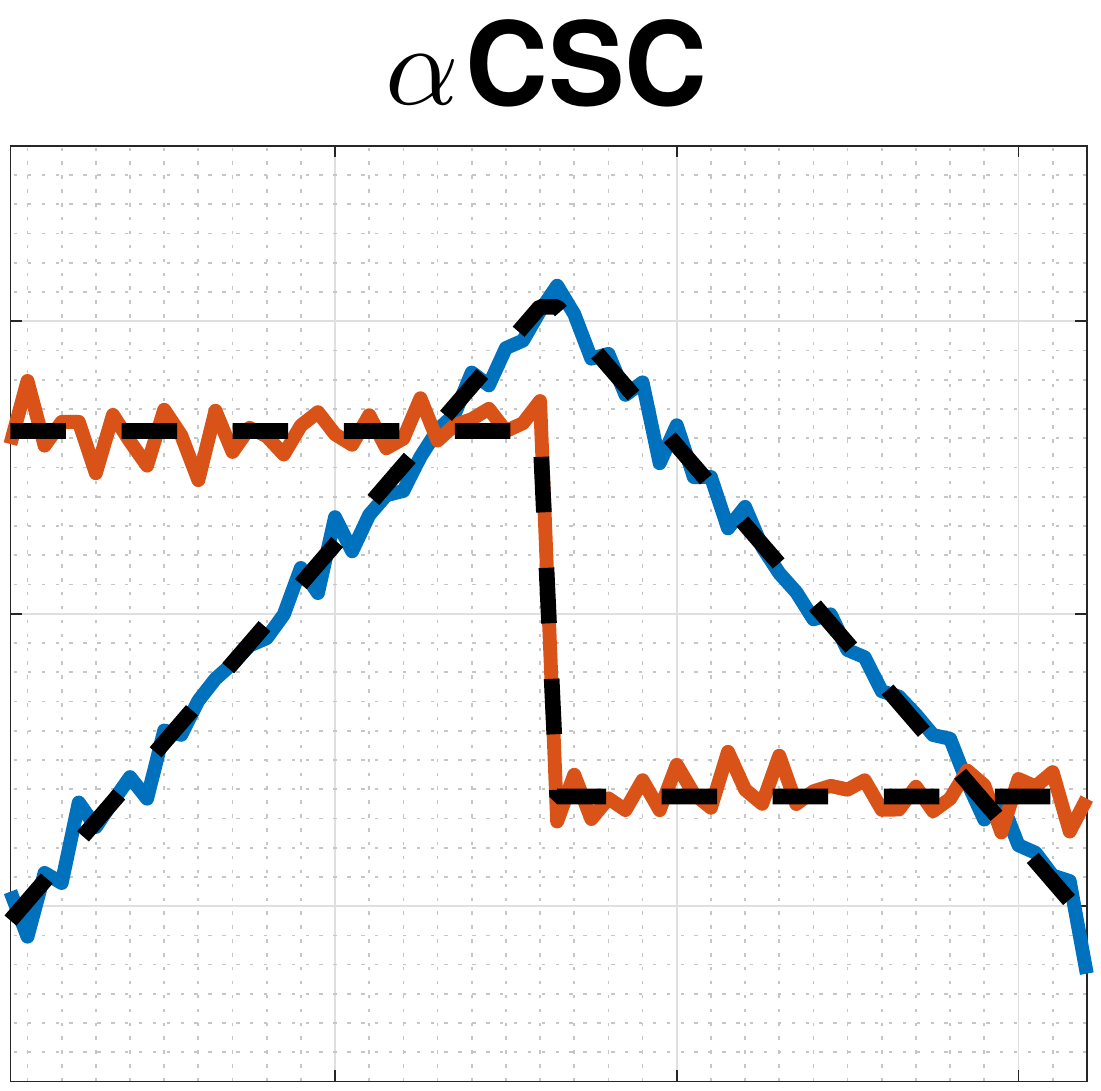}
        \end{tabular}
}\hfill
\subfigure[20\% corruption]{
\setlength{\tabcolsep}{0.5pt} 
\begin{tabular}{c c}
    \includegraphics[width=\tmpsize\linewidth]{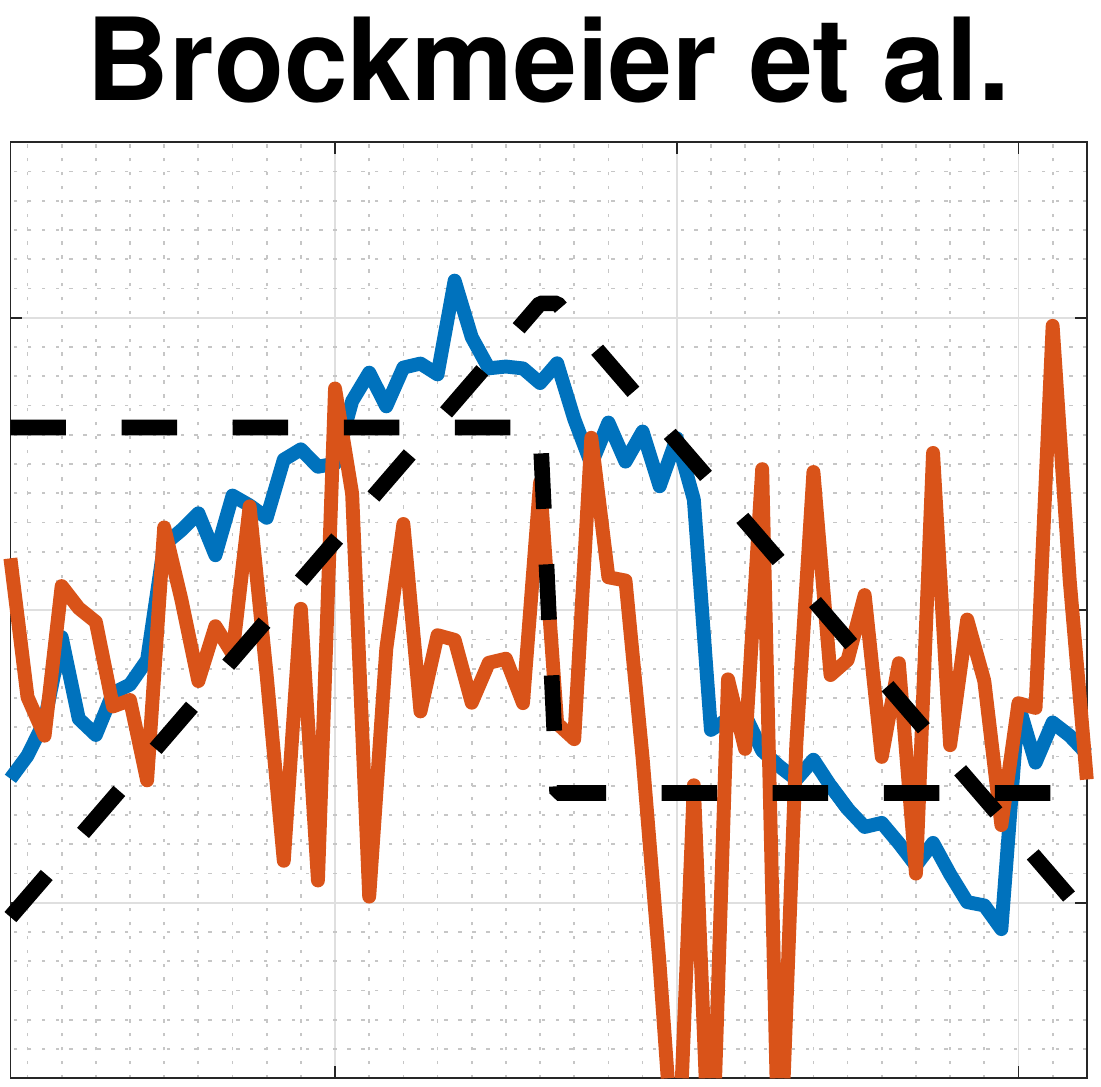} & 
    \includegraphics[width=\tmpsize\linewidth]{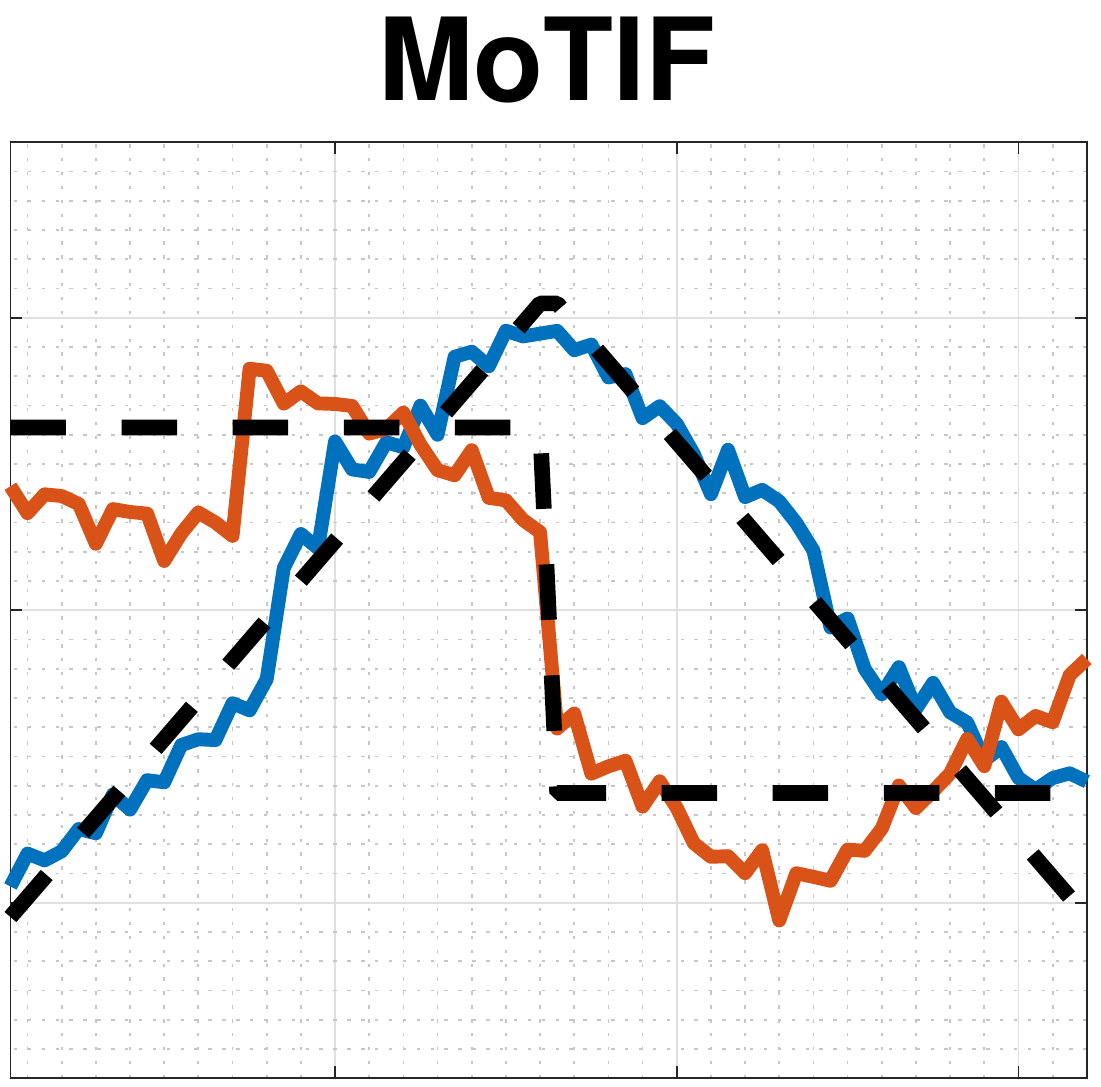}\\
    \includegraphics[width=\tmpsize\linewidth]{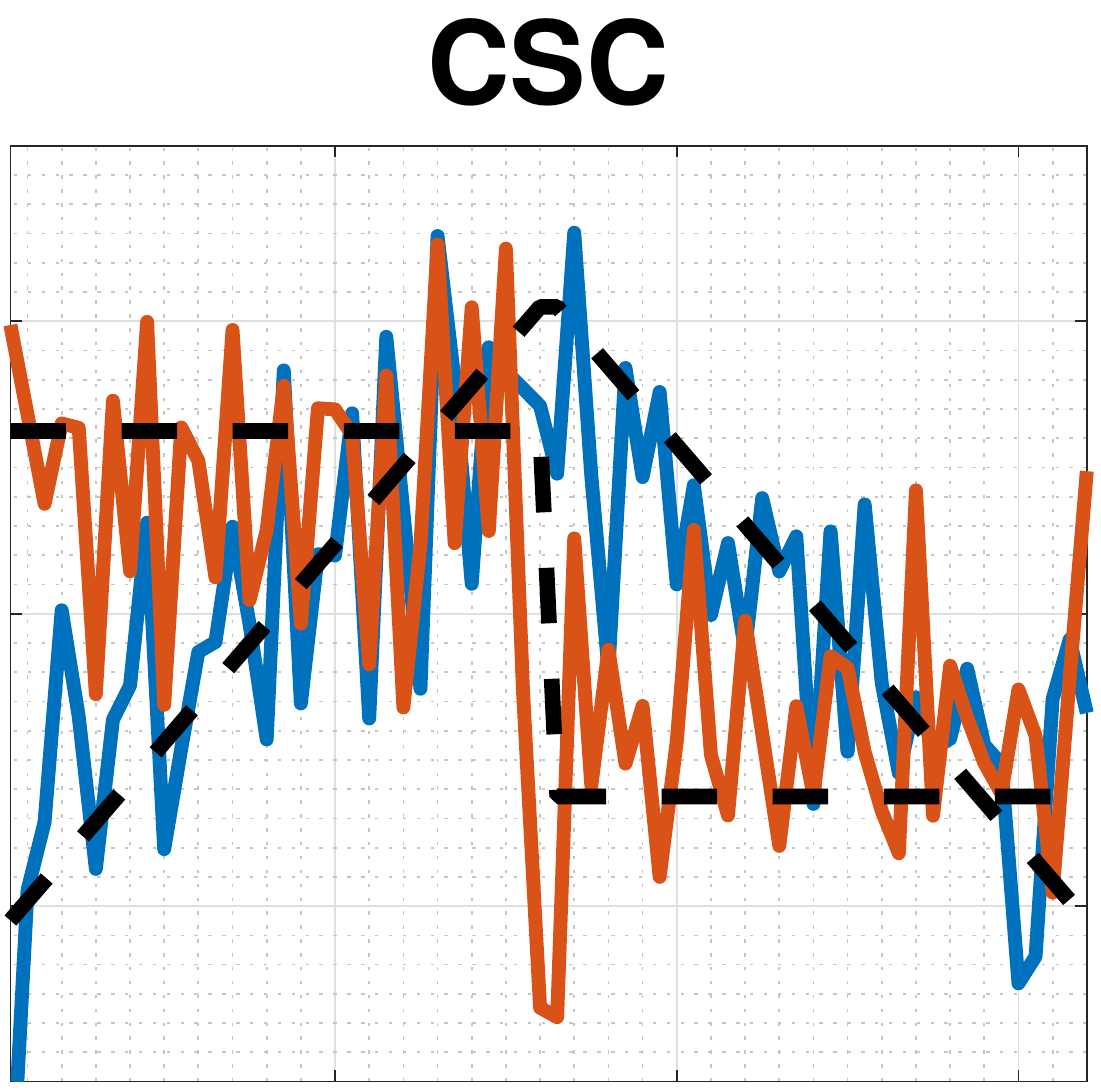}&
    \includegraphics[width=\tmpsize\linewidth]{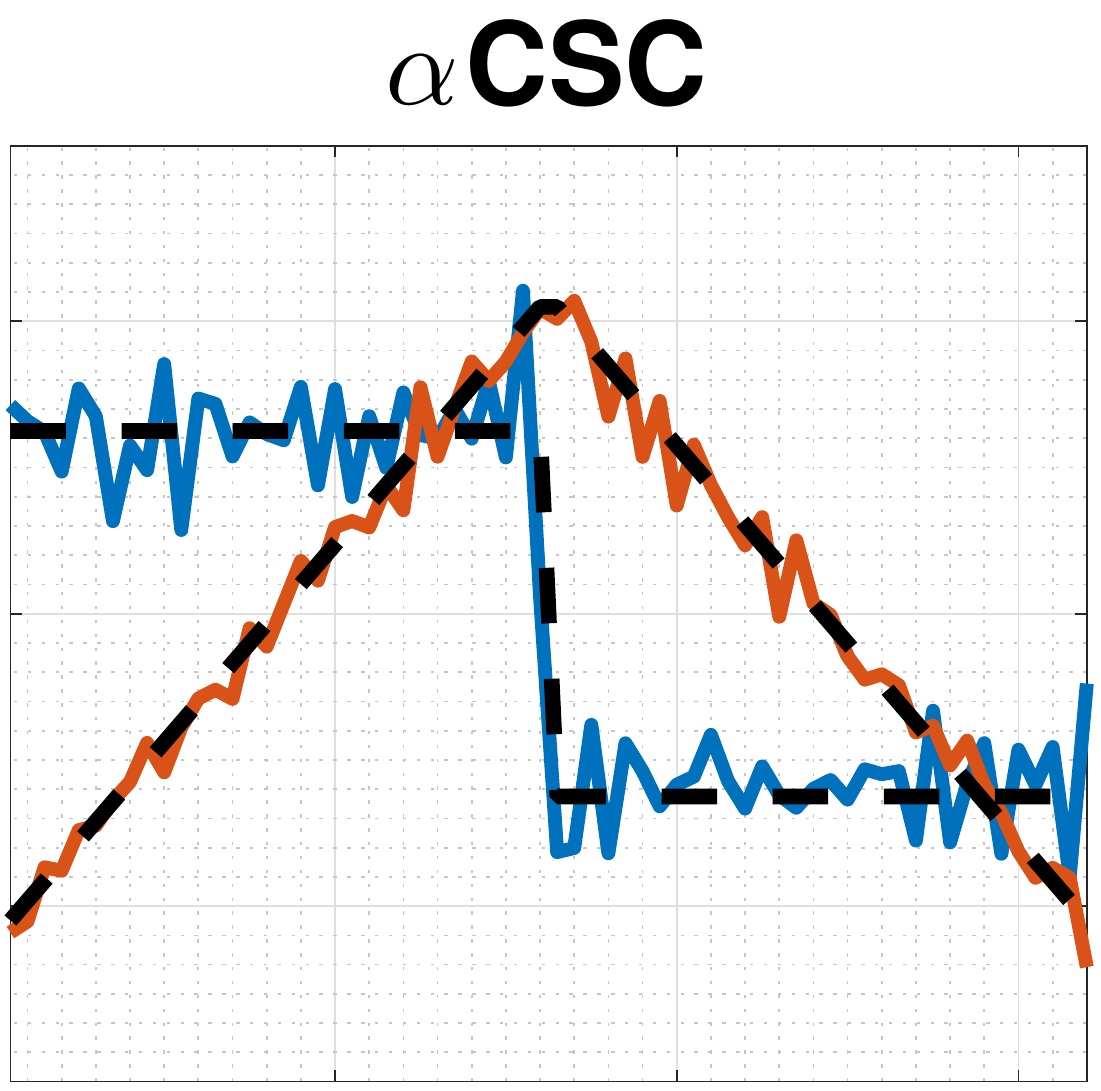}
    \end{tabular}
}
\end{center}
\vspace{-15pt}
\caption{Simulation to compare state-of-the-art methods against $\alpha$CSC.}
\label{fig:mcem_simulated} 
\vspace{-10pt}
\end{figure}

\textbf{Robustness to corrupted data:} 
In our second synthetic data experiment, we illustrate the robustness of $\alpha$CSC in the presence of corrupted observations.
In order to simulate the likely presence of high amplitude artifacts, one way would be to directly simulate the generative model in \eqref{eqn:acsc_org}. However, this would give us an unfair advantage, since $\alpha$CSC is specifically designed for such data. Here, we take an alternative approach, where we corrupt a randomly chosen fraction of the trials ($10\%$ or $20\%$) with strong Gaussian noise of standard deviation $0.1$, \textit{i.e.} one order of magnitude higher than in a regular trial. We used a regularization parameter of $\lambda = 0.1$.
In these experiments, by CSC we refer to $\alpha$CSC with $\alpha=2$, that resembles using only the M-step of our algorithm with deterministic weights $w_{n,t}=1/2$ for all $n$, $t$. We used a simpler setup where we set $N=100$, $T=512$, and $L=64$. We used $K=2$ atoms, as shown in dashed lines in Fig.~\ref{fig:mcem_simulated}.

For $\alpha$CSC, we set the number of outer iterations $I=5$, the number of iterations of the M-step to $M=50$, and the number of iterations of the MCMC algorithm to $J=10$. We discard the first $5$ samples of the MCMC algorithm as burn-in.
To enable a fair comparison, we run the standard CSC algorithm for $I\times M$ iterations, i.e.\ the \emph{total} number of M-step iterations in $\alpha$CSC. We also compared $\alpha$CSC against competing state-of-art methods previously applied to neural time series: \citet{brockmeier2016learning} and MoTIF~\cite{jost2006motif}. 
Starting from multiple random initializations, the estimated atoms with the smallest $\ell_2$ distance with the true atoms are shown in Fig.~\ref{fig:mcem_simulated}.

\begin{figure}[t]
    \centering
    \begin{minipage}{\linewidth}
        \begin{minipage}{0.75\textwidth}
             \subfigure[LFP spike data from \cite{hitziger2017adaptive}]{
             \includegraphics[height=2.3cm]{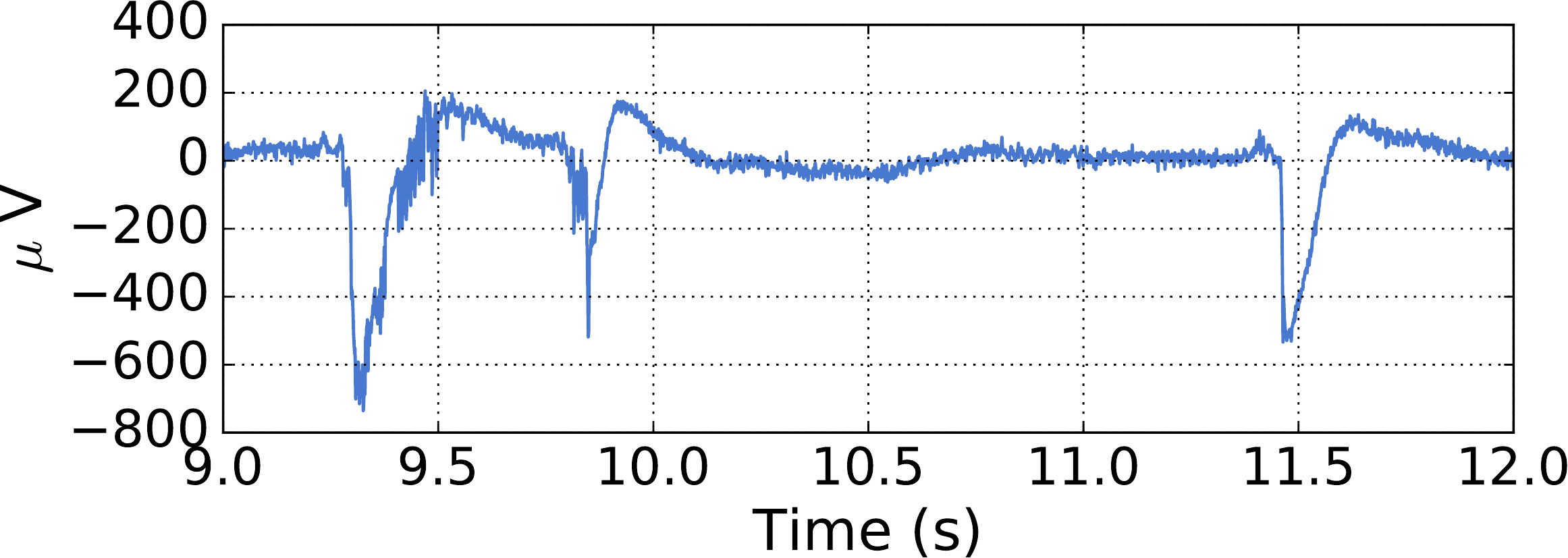}
             \label{fig:spikedata}}
             \subfigure[Estimated atoms]{
             \includegraphics[height=2.2cm]{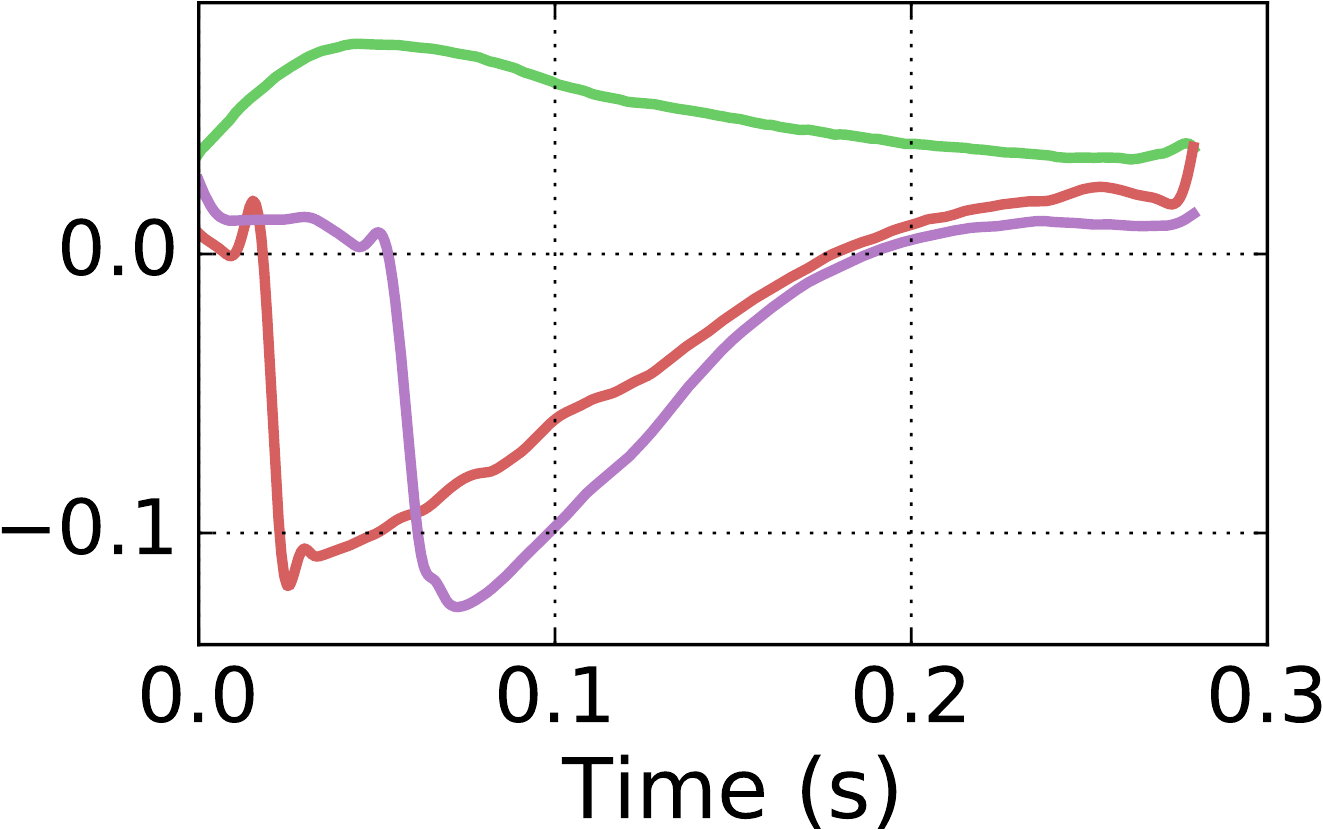}
             \label{fig:spikeatoms}}
        \end{minipage}
        \hfill
        \begin{minipage}{0.23\textwidth}
            \caption{Atoms learnt by $\alpha$CSC on LFP data containing epileptiform spikes with $\alpha=2$.}
        \end{minipage}
    \end{minipage}
\end{figure}

In the artifact-free scenario, all algorithms perform equally well, except for MoTIF that suffers from the presence of activations with varying amplitudes. This is because it aligns the data using correlations before performing the eigenvalue decomposition, without taking into account the strength of activations in each trial. The performance of \citet{brockmeier2016learning} and CSC degrades as the level of corruption increases. On the other hand, $\alpha$CSC is clearly more robust to the increasing level of corruption and recovers reasonable atoms even when 20\% of the trials are corrupted.

\textbf{Results on LFP data}
In our last set of experiments, we consider real neural data from two different datasets. 
We first applied $\alpha$CSC on an LFP dataset previously used in~\cite{hitziger2017adaptive} and containing epileptiform spikes as shown in Fig.~\ref{fig:spikedata}. The data was recorded in the rat cortex, and  is free of artifact. Therefore, we used the standard CSC with our optimization scheme, (i.e.\ $\alpha$CSC with $\alpha=2$).
As a standard preprocessing procedure, we applied a high-pass filter at $1$\,Hz in order to remove drifts in the signal, and then applied a tapered cosine window to down-weight the samples near the edges. We set $\lambda=6$, $N=300$, $T=2500$, $L=350$, and $K=3$. The recovered atoms by our algorithm are shown in Fig.~\ref{fig:spikeatoms}. We can observe that the estimated atoms resemble the spikes in Fig.~\ref{fig:spikedata}. These results show that, without using any heuristics, our approach can recover similar atoms to the ones reported in \cite{hitziger2017adaptive}, even though it does not make any assumptions on the shapes of the waveforms, or initializes the atoms with template spikes in order to ease the optimization.

The second dataset is an LFP channel in a rodent striatum from~\cite{dallerac2017updating}. We segmented the data into $70$ trials of length $2500$ samples, windowed each trial with a tapered cosine function, and detrended the data with a high-pass filter at $1$\,Hz.
We set $\lambda=10$, initialized the weights $w_n$ to the inverse of the variance of the trial $x_n$. Atoms are in all experiments initialized with Gaussian white noise.

\begin{figure}[t]
    \centering
    \subfigure[Atoms learnt by: CSC (clean data), CSC (full data), $\alpha$CSC (full data)]{
    \includegraphics[width=0.65\linewidth]{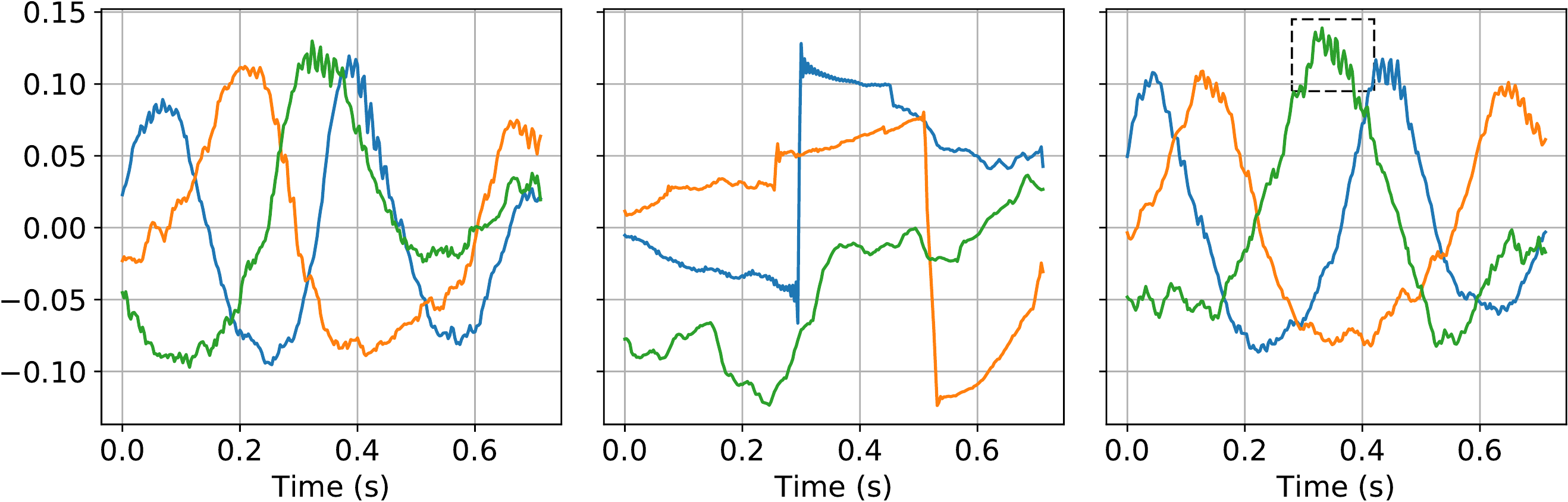}
    \label{fig:rat3atoms}
    }
    \subfigure[Comodulogram.]{
    \includegraphics[width=0.28\linewidth]{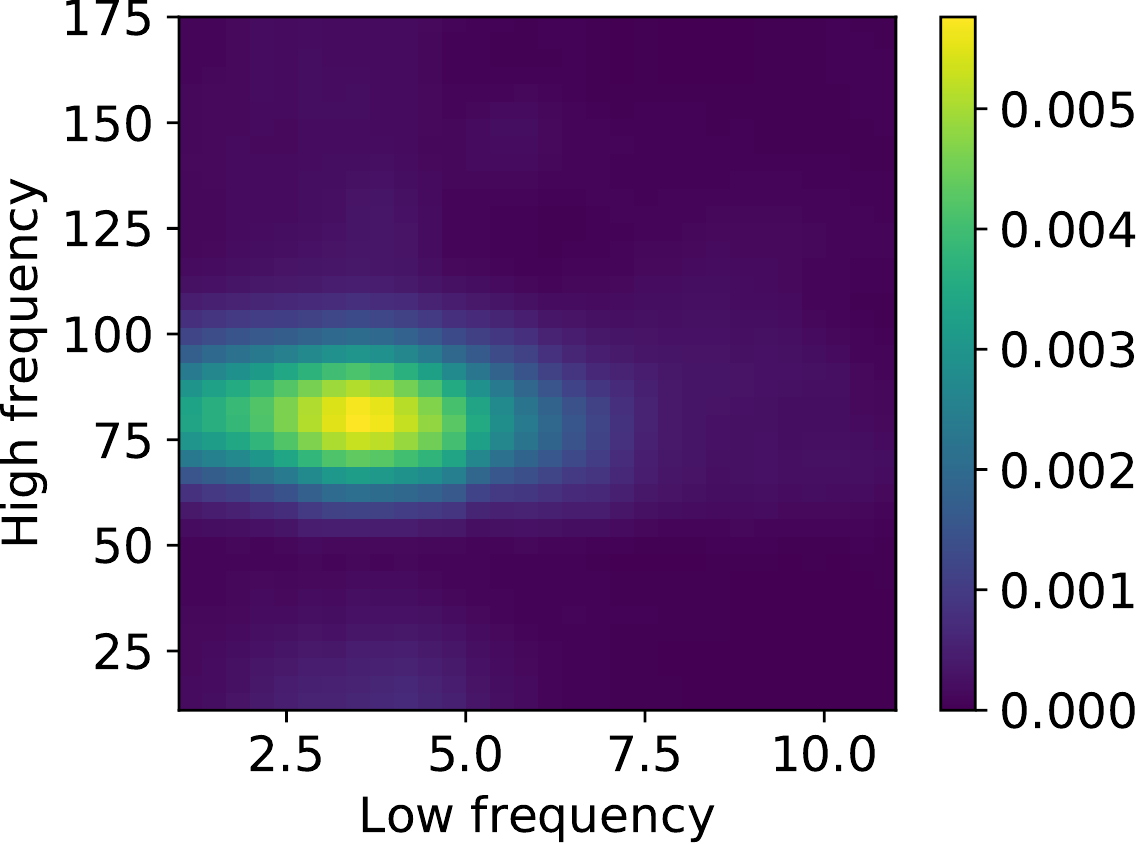}
    \label{fig:comodulogram}
    }
    \vspace{-10pt}
    \caption{(a)~Three atoms learnt from a rodent striatal LFP channel, using CSC on cleaned data, and both CSC and $\alpha$CSC on the full data. The atoms capture the cross-frequency coupling of the data (dashed rectangle). (b)~Comodulogram presents the cross-frequency coupling intensity computed between pairs of frequency bands on the entire cleaned signal, following \cite{tort2010measuring}.}
    \label{fig:ratdata}
    \vspace{-10pt}
\end{figure}

As opposed to the first LFP dataset, this dataset contains strong artifacts, as shown in Fig.~\ref{fig:artifacts}. In order to be able to illustrate the potential of CSC on this data, we first \emph{manually} identified and removed the trials that were corrupted by artifacts. In Fig.~\ref{fig:rat3atoms}, we illustrate the estimated atoms with CSC on the manually-cleaned data. We observe that the estimated atoms correspond to canonical waveforms found in the signal. In particular, the high frequency oscillations around $80$\,Hz are modulated in amplitude by the low-frequency oscillation around $3$\,Hz, a phenomenon known as cross-frequency coupling (CFC)~\cite{jensen2007cross}. We can observe this by computing a comodulogram~\cite{tort2010measuring} on the entire signal (Fig.~\ref{fig:comodulogram}). This measures the correlation between the amplitude of the high frequency band and the phase of the low frequency band.

Even though CSC is able to provide these excellent results on the cleaned data set, its performance heavily relies on the manual removal of the artifacts. Finally, we repeated the previous experiment on the full data, without removing the artifacts and compared CSC with $\alpha$CSC, where we set $\alpha=1.2$. The results are shown in the middle and the right sub-figures of Fig.~\ref{fig:rat3atoms}. It can be observed that in the presence of strong artifacts, CSC is not able to recover the atoms anymore. On the contrary, we observe that $\alpha$CSC can still recover atoms as observed in the artifact-free regime. In particular, the cross-frequency coupling phenomenon is still visible.

\section{Conclusion}
This work addresses the present need in the neuroscience community to better capture the complex morphology of brain waves~\cite{cole2017brain, gips2017discovering}. Our data-driven approach to this problem is a probabilistic formulation of a convolutional sparse coding model~\cite{Grosse-etal:2007}. We propose an inference strategy based on Monte Carlo EM to deal efficiently with heavy tailed noise and take into account the polarity of neural activations with a positivity constraint. Our problem formulation allows the use of fast quasi-Newton methods for the M-step which outperform previously proposed state-of-the-art ADMM-based algorithms~\cite{heide2015fast,wohlberg2016efficient,wohlberg2014efficient,bristow2013fast}, even when not making use of our parallel implementation. Results on LFP data demonstrate that such algorithms can be robust to the presence of transient artifacts in data and reveal insights on neural time-series without supervision.
We will make our code publicly available.

\section{Acknowledgement}
The work was supported by the French National Research Agency grants ANR-14-NEUC-0002-01 and ANR-16-CE23-0014 (FBIMATRIX), as well as the ERC Starting Grant SLAB ERC-YStG-676943.

\bibliographystyle{unsrtnat}

\bibliography{refs}

\begin{thebibliography}{35}
\providecommand{\natexlab}[1]{#1}
\providecommand{\url}[1]{\texttt{#1}}
\expandafter\ifx\csname urlstyle\endcsname\relax
  \providecommand{\doi}[1]{doi: #1}\else
  \providecommand{\doi}{doi: \begingroup \urlstyle{rm}\Url}\fi

\bibitem[Cole and Voytek(2017)]{cole2017brain}
S.~R. Cole and B.~Voytek.
\newblock Brain oscillations and the importance of waveform shape.
\newblock \emph{Trends Cogn. Sci.}, 2017.

\bibitem[Cohen(2014)]{cohen2014analyzing}
M.~X. Cohen.
\newblock \emph{Analyzing neural time series data: Theory and practice}.
\newblock MIT Press, 2014.
\newblock ISBN 9780262319560.

\bibitem[Jones(2016)]{jones2016brain}
S.~R. Jones.
\newblock When brain rhythms aren't ‘rhythmic’: implication for their
  mechanisms and meaning.
\newblock \emph{Curr. Opin. Neurobiol.}, 40:\penalty0 72--80, 2016.

\bibitem[Mazaheri and Jensen(2008)]{mazaheri2008asymmetric}
A.~Mazaheri and O.~Jensen.
\newblock Asymmetric amplitude modulations of brain oscillations generate slow
  evoked responses.
\newblock \emph{{The Journal of Neuroscience}}, 28\penalty0 (31):\penalty0
  7781--7787, 2008.

\bibitem[Hari and Puce(2017)]{hari2017meg}
R.~Hari and A.~Puce.
\newblock \emph{{MEG-EEG Primer}}.
\newblock Oxford University Press, 2017.

\bibitem[Jost et~al.(2006)Jost, Vandergheynst, Lesage, and
  Gribonval]{jost2006motif}
P.~Jost, P.~Vandergheynst, S.~Lesage, and R.~Gribonval.
\newblock {MoTIF: an efficient algorithm for learning translation invariant
  dictionaries}.
\newblock In \emph{{Acoustics, Speech and Signal Processing, ICASSP}},
  volume~5. IEEE, 2006.

\bibitem[Brockmeier and Pr{\'\i}ncipe(2016)]{brockmeier2016learning}
A.~J. Brockmeier and J.~C. Pr{\'\i}ncipe.
\newblock Learning recurrent waveforms within {EEGs}.
\newblock \emph{IEEE Transactions on Biomedical Engineering}, 63\penalty0
  (1):\penalty0 43--54, 2016.

\bibitem[Hitziger et~al.(2017)Hitziger, Clerc, Saillet, Benar, and
  Papadopoulo]{hitziger2017adaptive}
S.~Hitziger, M.~Clerc, S.~Saillet, C.~Benar, and T.~Papadopoulo.
\newblock {Adaptive Waveform Learning: A Framework for Modeling Variability in
  Neurophysiological Signals}.
\newblock \emph{IEEE Transactions on Signal Processing}, 2017.

\bibitem[Gips et~al.(2017)Gips, Bahramisharif, Lowet, Roberts, de~Weerd,
  Jensen, and van~der Eerden]{gips2017discovering}
B.~Gips, A.~Bahramisharif, E.~Lowet, M.~Roberts, P.~de~Weerd, O.~Jensen, and
  J.~van~der Eerden.
\newblock Discovering recurring patterns in electrophysiological recordings.
\newblock \emph{{J. Neurosci. Methods}}, 275:\penalty0 66--79, 2017.

\bibitem[Grosse et~al.(2007)Grosse, Raina, Kwong, and Ng]{Grosse-etal:2007}
R.~Grosse, R.~Raina, H.~Kwong, and A.~Y. Ng.
\newblock Shift-invariant sparse coding for audio classification.
\newblock In \emph{23rd Conference on Uncertainty in Artificial Intelligence},
  UAI'07, pages 149--158. AUAI Press, 2007.
\newblock ISBN 0-9749039-3-0.

\bibitem[Heide et~al.(2015)Heide, Heidrich, and Wetzstein]{heide2015fast}
F.~Heide, W.~Heidrich, and G.~Wetzstein.
\newblock Fast and flexible convolutional sparse coding.
\newblock In \emph{{Computer Vision and Pattern Recognition (CVPR)}}, pages
  5135--5143. IEEE, 2015.

\bibitem[Wohlberg(2016)]{wohlberg2016efficient}
B.~Wohlberg.
\newblock {Efficient algorithms for convolutional sparse representations}.
\newblock \emph{{Image Processing, IEEE Transactions on}}, 25\penalty0
  (1):\penalty0 301--315, 2016.

\bibitem[Zeiler et~al.(2010)Zeiler, Krishnan, Taylor, and
  Fergus]{zeiler2010deconvolutional}
M.~D. Zeiler, D.~Krishnan, G.W. Taylor, and R.~Fergus.
\newblock Deconvolutional networks.
\newblock In \emph{{Computer Vision and Pattern Recognition (CVPR)}}, pages
  2528--2535. IEEE, 2010.

\bibitem[{\v{S}}orel and {\v{S}}roubek(2016)]{vsorel2016fast}
M.~{\v{S}}orel and F.~{\v{S}}roubek.
\newblock Fast convolutional sparse coding using matrix inversion lemma.
\newblock \emph{{Digital Signal Processing}}, 2016.

\bibitem[Kavukcuoglu et~al.(2010)Kavukcuoglu, Sermanet, Boureau, Gregor,
  Mathieu, and Cun]{kavukcuoglu2010learning}
K.~Kavukcuoglu, P.~Sermanet, Y-L. Boureau, K.~Gregor, M.~Mathieu, and Y.~Cun.
\newblock Learning convolutional feature hierarchies for visual recognition.
\newblock In \emph{Advances in Neural Information Processing Systems (NIPS)},
  pages 1090--1098, 2010.

\bibitem[Pachitariu et~al.(2013)Pachitariu, Packer, Pettit, Dalgleish, Hausser,
  and Sahani]{pachitariu2013extracting}
M.~Pachitariu, A.~M Packer, N.~Pettit, H.~Dalgleish, M.~Hausser, and M.~Sahani.
\newblock Extracting regions of interest from biological images with
  convolutional sparse block coding.
\newblock In \emph{Advances in Neural Information Processing Systems (NIPS)},
  pages 1745--1753, 2013.

\bibitem[Mailh{\'e} et~al.(2008)Mailh{\'e}, Lesage, Gribonval, Bimbot, and
  Vandergheynst]{mailhe2008shift}
B.~Mailh{\'e}, S.~Lesage, R.~Gribonval, F.~Bimbot, and P.~Vandergheynst.
\newblock Shift-invariant dictionary learning for sparse representations:
  extending {K-SVD}.
\newblock In \emph{16th Eur. Signal Process. Conf.}, pages 1--5. IEEE, 2008.

\bibitem[Barth{\'e}lemy et~al.(2013)Barth{\'e}lemy, Gouy-Pailler, Isaac,
  Souloumiac, Larue, and Mars]{barthelemy2013multivariate}
Q.~Barth{\'e}lemy, C.~Gouy-Pailler, Y.~Isaac, A.~Souloumiac, A.~Larue, and
  J.~I. Mars.
\newblock {Multivariate temporal dictionary learning for {EEG}}.
\newblock \emph{{J. Neurosci. Methods}}, 215\penalty0 (1):\penalty0 19--28,
  2013.

\bibitem[Samorodnitsky and Taqqu(1994)]{samorodnitsky1994stable}
G.~Samorodnitsky and M.~S. Taqqu.
\newblock \emph{Stable non-{G}aussian random processes: stochastic models with
  infinite variance}, volume~1.
\newblock CRC press, 1994.

\bibitem[Kuruoglu(1999)]{kuruoglu1999signal}
E.~E. Kuruoglu.
\newblock \emph{Signal processing in $\alpha$-stable noise environments: a
  least {L}p-norm approach}.
\newblock PhD thesis, University of Cambridge, 1999.

\bibitem[Mandelbrot(2013)]{mandelbrot2013fractals}
B.~B. Mandelbrot.
\newblock \emph{Fractals and scaling in finance: Discontinuity, concentration,
  risk. Selecta volume E}.
\newblock Springer Science \& Business Media, 2013.

\bibitem[Wang et~al.(2016)Wang, Qi, Wang, Lei, Zheng, and Pan]{wang2016delving}
Y.~Wang, Y.~Qi, Y.~Wang, Z.~Lei, X.~Zheng, and G.~Pan.
\newblock Delving into $\alpha$-stable distribution in noise suppression for
  seizure detection from scalp {EEG}.
\newblock \emph{J. Neural. Eng.}, 13\penalty0 (5):\penalty0 056009, 2016.

\bibitem[Chambers et~al.(1976)Chambers, Mallows, and Stuck]{chambers1976method}
J.~M. Chambers, C.~L. Mallows, and B.~W. Stuck.
\newblock A method for simulating stable random variables.
\newblock \emph{Journal of the american statistical association}, 71\penalty0
  (354):\penalty0 340--344, 1976.

\bibitem[Huber(1981)]{Huber81a}
P.~J. Huber.
\newblock \emph{Robust Statistics}.
\newblock Wiley, 1981.

\bibitem[Godsill and Kuruoglu(1999)]{godsill1999bayesian}
S.~Godsill and E.~Kuruoglu.
\newblock Bayesian inference for time series with heavy-tailed symmetric
  $\alpha$-stable noise processes.
\newblock \emph{Proc. Applications of heavy tailed distributions in economics,
  eng. and stat.}, 1999.

\bibitem[Chib and Greenberg(1995)]{chib1995understanding}
S.~Chib and E.~Greenberg.
\newblock Understanding the {M}etropolis-{H}astings algorithm.
\newblock \emph{{The American Statistician}}, 49\penalty0 (4):\penalty0
  327--335, 1995.

\bibitem[Liu(2008)]{Liu2008}
J.S. Liu.
\newblock \emph{Monte Carlo strategies in scientific computing}.
\newblock Springer, 2008.

\bibitem[Byrd et~al.(1995)Byrd, Lu, Nocedal, and Zhu]{byrd1995limited}
R.~H. Byrd, P.~Lu, J.~Nocedal, and C.~Zhu.
\newblock A limited memory algorithm for bound constrained optimization.
\newblock \emph{SIAM Journal on Scientific Computing}, 16\penalty0
  (5):\penalty0 1190--1208, 1995.

\bibitem[Moulines et~al.(1995)Moulines, Duhamel, Cardoso, and
  Mayrargue]{moulines1995subspace}
E.~Moulines, P.~Duhamel, J-F. Cardoso, and S.~Mayrargue.
\newblock Subspace methods for the blind identification of multichannel {FIR}
  filters.
\newblock \emph{IEEE Transactions on signal processing}, 43\penalty0
  (2):\penalty0 516--525, 1995.

\bibitem[Beck and Teboulle(2009)]{beck2009fast}
A.~Beck and M.~Teboulle.
\newblock A fast iterative shrinkage-thresholding algorithm for linear inverse
  problems.
\newblock \emph{SIAM journal on imaging sciences}, 2\penalty0 (1):\penalty0
  183--202, 2009.

\bibitem[Dall{\'e}rac et~al.(2017)Dall{\'e}rac, Graupner, Knippenberg,
  Martinez, Tavares, Tallot, El~Massioui, Verschueren, H{\"o}hn, Bertolus,
  et~al.]{dallerac2017updating}
G.~Dall{\'e}rac, M.~Graupner, J.~Knippenberg, R.~C.~R. Martinez, T.~F. Tavares,
  L.~Tallot, N.~El~Massioui, A.~Verschueren, S.~H{\"o}hn, J.B. Bertolus, et~al.
\newblock Updating temporal expectancy of an aversive event engages striatal
  plasticity under amygdala control.
\newblock \emph{Nature Communications}, 8:\penalty0 13920, 2017.

\bibitem[Jensen and Colgin(2007)]{jensen2007cross}
O.~Jensen and L.~L. Colgin.
\newblock {Cross-frequency coupling between neuronal oscillations}.
\newblock \emph{{Trends in cognitive sciences}}, 11\penalty0 (7):\penalty0
  267--269, 2007.

\bibitem[Tort et~al.(2010)Tort, Komorowski, Eichenbaum, and
  Kopell]{tort2010measuring}
A.~BL. Tort, R.~Komorowski, H.~Eichenbaum, and N.~Kopell.
\newblock Measuring phase-amplitude coupling between neuronal oscillations of
  different frequencies.
\newblock \emph{J. Neurophysiol.}, 104\penalty0 (2):\penalty0 1195--1210, 2010.

\bibitem[Wohlberg(2014)]{wohlberg2014efficient}
B.~Wohlberg.
\newblock Efficient convolutional sparse coding.
\newblock In \emph{{Acoustics, Speech and Signal Processing, ICASSP}}, pages
  7173--7177. IEEE, 2014.

\bibitem[Bristow et~al.(2013)Bristow, Eriksson, and Lucey]{bristow2013fast}
H.~Bristow, A.~Eriksson, and S.~Lucey.
\newblock Fast convolutional sparse coding.
\newblock In \emph{{Computer Vision and Pattern Recognition (CVPR)}}, pages
  391--398, 2013.

\end{thebibliography}


\begin{thebibliography}{2}
\providecommand{\natexlab}[1]{#1}
\providecommand{\url}[1]{\texttt{#1}}
\expandafter\ifx\csname urlstyle\endcsname\relax
  \providecommand{\doi}[1]{doi: #1}\else
  \providecommand{\doi}{doi: \begingroup \urlstyle{rm}\Url}\fi

\bibitem[Mairal et~al.(2010)Mairal, Bach, Ponce, and Sapiro]{mairal2010online}
J.~Mairal, F.~Bach, J.~Ponce, and G.~Sapiro.
\newblock Online learning for matrix factorization and sparse coding.
\newblock \emph{Journal of Machine Learning Research}, 11\penalty0
  (Jan):\penalty0 19--60, 2010.

\bibitem[Boyd et~al.(2011)Boyd, Parikh, Chu, Peleato, and
  Eckstein]{boyd2011distributed}
S.~Boyd, N.~Parikh, E.~Chu, B.~Peleato, and J.~Eckstein.
\newblock Distributed optimization and statistical learning via the alternating
  direction method of multipliers.
\newblock \emph{Foundations and Trends{\textregistered} in Machine Learning},
  3\penalty0 (1):\penalty0 1--122, 2011.

\end{thebibliography}

\newpage
\appendix

\section{Details of the E-Step}

Computing the weights that are required in the M-step requires us to compute the expectation of $\frac1{\phi_{n,t}}$ under the posterior distribution $p(\phi_{n,t}|x,d,z)$, which is not analytically available. 

Monte Carlo methods are numerical techniques that can be used to approximately compute the expectations of the form:
\begin{align}
\mathds{E}[f(\phi_{n,t})] = \int f(\phi_{n,t}) \pi(\phi_{n,t}) d\phi_{n,t} \approx \frac1{J} \sum_{j=1}^J f(\phi_{n,t}^{(j)}) \label{eqn:mc}
\end{align}
where $\phi_{n,t}^{(j)}$ are some samples drawn from $\pi(\phi_{n,t}) \triangleq p(\phi_{n,t}|x,d,z)$ and $f(\phi) = 1/\phi$ in our case. However, in our case, sampling directly from $\pi(\phi_{n,t})$ is also unfortunately intractable.

MCMC methods generate samples from the target distribution $\pi(\phi_{n,t})$ by forming a Markov chain, whose stationary distribution is $\pi(\phi_{n,t})$, 
so that $\pi(\phi_{n,t}) = \int {\cal T}(\phi_{n,t}|\phi_{n,t}') p(\phi_{n,t}') d\phi_{n,t}'$, where ${\cal T}$ denotes the transition kernel of the Markov chain. 

In this study, we develop a Metropolis-Hastings (MH) algorithm, that implicitly forms a transition kernel. 
The MH algorithm generates samples from a target distribution $\pi(\phi_{n,t})$ in two steps. First, it generates a random sample $\phi_{n,t}'$ from a \emph{proposal} distribution $\phi_{n,t}' \sim q(\phi_{n,t}'|\phi_{n,t}^{(j)})$, then computes an acceptance probability $\text{acc}(\phi_{n,t}^{(j)} \rightarrow \phi_{n,t}')$ and draws a uniform random number $u \sim {\cal U}([0, 1])$. If $u < \text{acc}(\phi_{n,t}^{(j)} \rightarrow \phi_{n,t}')$, it accepts the sample and sets $\phi_{n,t}^{(j+1)} = \phi_{n,t}'$; otherwise it rejects the sample and sets $\phi_{n,t}^{(j+1)} = \phi_{n,t}^{(j)}$. The acceptance probability is given as follows
\begin{align}
\text{acc}(\phi_{n,t} \rightarrow \phi_{n,t}') = \min \Bigr\{1, \frac{q(\phi_{n,t}|\phi_{n,t}') \pi(\phi_{n,t}')}{q(\phi_{n,t}'|\phi_{n,t}) \pi(\phi_{n,t})}\Bigr\} = \min \Bigr\{1, \frac{q(\phi_{n,t}|\phi_{n,t}') p(x_{n,t}|\phi_{n,t}',d,z) p(\phi_{n,t}') }{q(\phi_{n,t}'|\phi_{n,t}) p(x_{n,t}|\phi_{n,t},d,z) p(\phi_{n,t}) }\Bigr\}
\end{align}
where the last equality is obtained by applying the Bayes rule on $\pi$. 

The acceptance probability requires the prior distribution of $\phi$ to be evaluated. Unfortunately, this is intractable in our case since this prior distribution is chosen to be a positive $\alpha$-stable distribution whose PDF does not have an analytical form. As a remedy, we choose the prior distribution of $\phi_{n,t}$ as the proposal distribution, such $q(\phi_{n,t}|\phi_{n,t}') = p(\phi_{n,t})$. This enables us to simplify the acceptance probability. Accordingly, for each $\phi_{n,t}$, we have the following acceptance probability:
\begin{align}
  \text{acc}(\phi_{n,t}^{(i,j)} \rightarrow \phi_{n,t}' ) \triangleq \min  \Bigl\{1, \exp(\log \phi_{n,t}^{(i,j)} - \log \phi_{n,t}')/2 + (x_{n,t} - \hat{x}^{(i)}_{n,t})^2 (1/{\phi_{n,t}^{(i,j)}} - 1/{\phi_{n,t}'}) \Bigr\}.
\end{align}
Thanks to the simplification, this probability is tractable and can be easily computed. 

\section{Details of the M-Step}

\subsection{Solving for the activations}
In the M-step, we optimize~\eqref{eq:problem_definition_z} to find the activations $z_n^{(i)}$ of each trial $n$ independently. To keep the notation simple, we will drop the index for the iteration number $i$ of the EM algorithm.

First, this equation can be rewritten by concatenating the Toeplitz matrices for the $K$ atoms into a big matrix $D = [D^1, D^2, ..., D^K] \in \bbR^{T \times KT}$ and the activations for different atoms into a single vector $\bar{z}_n = [(\bar{z}_n^1)^\top, (\bar{z}_n^2)^\top, ..., (\bar{z}_n^K)^\top]^\top \in \bbR^{KT}_+$ where $(\cdot)^\top$ denotes the transposition operation. Recall that $\bar{z}_n^k$ is a zero-padded version of $z_n^k$. This leads to a simpler formulation and the objective function $\mathcal{L}(d, z)$:
\begin{equation}
\mathcal{L}(d, z) = \sum_{n=1}^{N} \frac{1}{2}\|\sqrt{w_{n}} \odot (x_{n} - D \bar{z}_{n})\|_{2}^{2} + \lambda \mathbbm{1}^\top \bar{z}_{n} \enspace ,
\label{eq:problem_definition_d_simple}
\end{equation}
where $\mathbbm{1} \in \bbR^{KT}$ is a vector of ones.

The derivative w.r.t. $z_n$ now reads:
\begin{equation}
\frac{\partial \mathcal{L}(d, z)}{\partial \bar{z}_n}
= D^\top(w_n \odot (x_n - D\bar{z}_n)) + \lambda \mathbbm{1}^\top \enspace .
\end{equation}
In practice, this big matrix $D$ is never assembled and all operations are carried out using convolutions. Note also that we do not update the zeros from the padding in $\bar{z}_n^k$. Now that we have the gradient, the activations can be estimated using a efficient quasi-Newton solver such as L-BFGS-B, taking into account the box posititivy constraint $0 \leq z_n \leq \infty$.

For each trial, one iteration costs $\mathcal{O}(LKT)$.

\subsection{Solving for the atoms}

In the M-step, we optimize \eqref{eq:problem_definition_d} to find the atoms $d^k$.
As when solving for the activations $z_n$, we can remove the summation over the atoms by concatenating the delayed matrices into $Z_{n}=[Z_{n}^1, Z_{n}^2, \dots, Z_{n}^K] \in \bbR^{T \times KL}$ and $d=[(d^1)^\top, (d^2)^\top, ..., (d^K)^\top]^\top \in \bbR^{KL}$. This leads to the simpler formulation:
\begin{align}
& \min_{d} \sum_{n=1}^{N} \frac{1}{2}\|\sqrt{w_n} \odot (x_{n} - Z_{n}d)\|_{2}^{2}, \quad \text{  s.t.  } \|d^k\|_2^2 \leq 1 \enspace.
\label{eq:subproblem_d}
\end{align}
 The Lagrangian of this problem is given by:
\begin{equation}
g(d, \beta) = \sum_{n=1}^{N} \frac{1}{2}\|\sqrt{w_n} \odot (x_{n} - \sum_{k=1}^{K} Z_{n}^{k}d^{k}) \|_{2}^{2} + \sum_k \beta^k (\|d^k\|_2^{2} - 1) \quad \text{s.t. } \beta^k \geq 0 \enspace,
\end{equation}
where $\beta = (\beta^1, \beta^2, ..., \beta^K)$ are the dual variables. Therefore, the dual problem is:
\begin{align}
\min_{d}{g(d, \beta)} = g(d^{*}, \beta)
\end{align}
where $d^*$, the primal optimal, is given by:
\begin{equation}
d^{*} = (\sum_{n=1}^N Z_n^{\top}(w_{n} \odot Z_n) + \bar{\beta} )^{-1}\sum_{n=1}^{N}(w_{n} \odot Z_n)^{\top}x_n
\label{eq:dual_optimal}
\end{equation}
with $\bar{\beta} = \mathrm{diag}([\mathbbm{1}\beta^1, \mathbbm{1}\beta^2, ..., \mathbbm{1}\beta^K]) \in \bbR^{KL}$ with $\mathbbm{1} \in \bbR^{L}$. The gradient for the dual variable $\beta^k$ is given by:
\begin{equation}
\frac{\partial g(d^{*}, \beta)}{\partial \beta^k}  = \|{d^{*}}^k\|_2^2 - 1,
\end{equation}
with ${d^{*}}^k$ computed from~\eqref{eq:dual_optimal}.
We can solve this iteratively using again L-BFGS-B taking into account the positivity constraint $\beta^k \geq 0$ for all $k$.
What we have described so far solves for all the atoms simultaneously. However, it is also possible to estimate the atoms sequentially one at a time using a block coordinate descent (BCD) approach, as in the work of \citelatex{mairal2010online}. In each iteration of the BCD algorithm, a residual $r_n^k$ is computed as given by:
\begin{equation}
r_n^k = x_n - \sum_{k'\neq k} Z^{k'}_{n}d^{k'}
\end{equation}
and correspondingly subproblem \ref{eq:subproblem_d} becomes:
\begin{align}
& \min_{d^k} \sum_{n=1}^{N} \frac{1}{2}\|\sqrt{w_n} \odot (r^k_{n} - Z^k_{n}d^k)\|_{2}^{2}, \quad \text{  s.t.  } \|d^k\|_2^2 \leq 1, \enspace.
\label{eq:subproblem_d_block}
\end{align}
which is solved in the same way as subproblem~\ref{eq:subproblem_d}. Now, in the simultaneous case, we construct one linear problem in $\mathcal{O}(L^2K^2TN)$ and one iteration costs $\mathcal{O}(L^3K^3)$. However, in the BCD strategy, we construct $K$ linear problems in $\mathcal{O}(L^2TN)$ and one iteration costs only $\mathcal{O}(L^3)$.
Interestingly, when the weights $w_n$ are all identical, we can 
use the fact that for one atom $k$, the matrix $\sum_{i=1}^{N}(Z_i^k)^T Z_i^k$ is Toeplitz. In this case, we can construct $K$ linear problems in only $\mathcal{O}(LTN)$ and one iteration costs only $\mathcal{O}(L^2)$.

For the benefit of the reader, we summarize the complexity of the M-step in Table~\ref{table:complexity_m}. We note $p$ and $q$ the number of iterations in the L-BFGS-B methods for the activations update and atoms update.

\begin{table}[htb]
\begin{center}
\begin{tabular}{|l|l|}
\hline
Method & Complexity \\
\hline
Solving activations $z$ & $p\min(L, \log(T))KTN$ \\
Solving atoms $d$ & $L^2K^2TN + qL^3K^3$ \\
Solving atoms $d$ (sequential) & $LKTN + qL^2K$ \\
\hline
\end{tabular}
\vspace{5pt}
\caption{Complexity analysis of the M-step, where $p$ and $q$ are the number of iterations in the L-BFGS-B solvers for the activations and atoms updates.}
\label{table:complexity_m}
\end{center}
\end{table}

\section{Additional Experiments: M-step speed benchmark}

\subsection{Comparison with state of the art}

\begin{figure}[htb]
    \centering
     \subfigure[$K=2$, $L=32$.]{
     \includegraphics[width=0.32\linewidth]{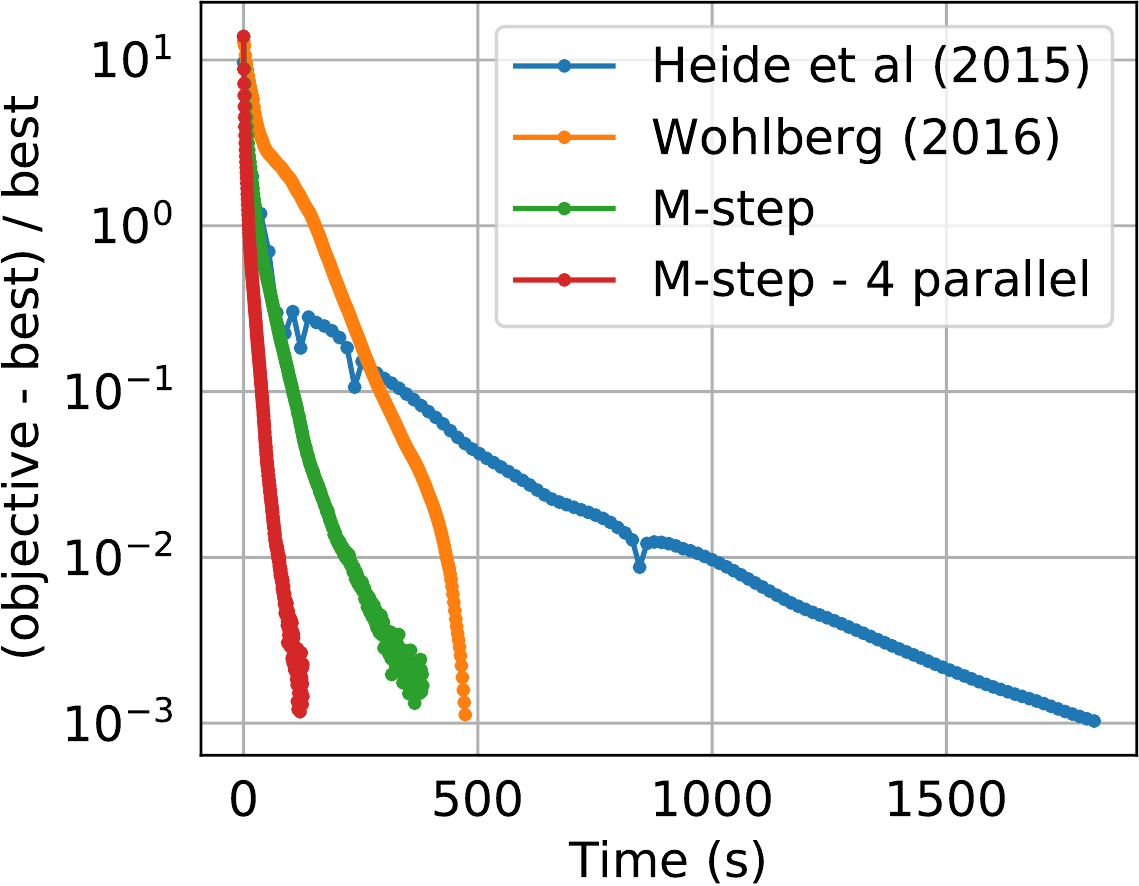}}
     \subfigure[$K=2$, $L=128$.]{
     \includegraphics[width=0.32\textwidth]{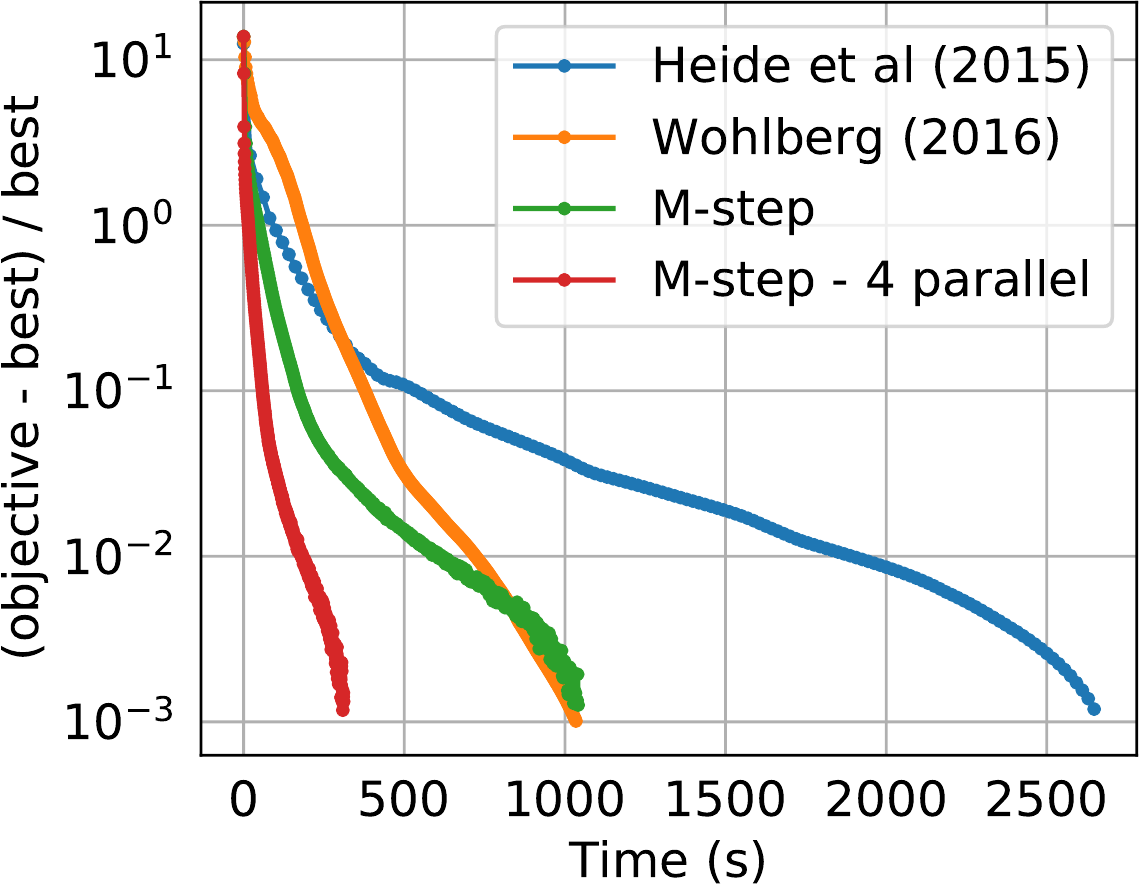}}
     \subfigure[$K=10$, $L=32$.]{
     \includegraphics[width=0.32\textwidth]{figures/relative_10_32.pdf}}
    \vspace{-5pt}
    \caption{Convergence speed of the relative objective function. The y-axis shows the objective function relative to the obtained minimum for each run: $(f(x) - f(x^*))/f(x^*)$. Each curve is the geometrical mean over 24 different random initializations.}
    \label{fig:convergence_setups}
\end{figure}

Here, we compare convergence plots of our algorithm against a number of state-of-art methods. The details of the experimental setup are described in Section~\ref{sec:experiments}. Fig.~\ref{fig:convergence_setups} demonstrates on a variety of setups the computational
advantage of our quasi-Newton approach to solve the M-step.
Note that Fig.~\ref{fig:convergence}b is in fact a summary of Fig.~\ref{fig:convergence_setups}. 
Indeed, we can verify that ADMM methods converge quickly to a modest accuracy, but take much longer to converge to a high accuracy.\citelatex{boyd2011distributed}

\begin{figure}[htb]
    \centering
     \subfigure[$K=2$, $L=32$.]{
     \includegraphics[width=0.32\linewidth]{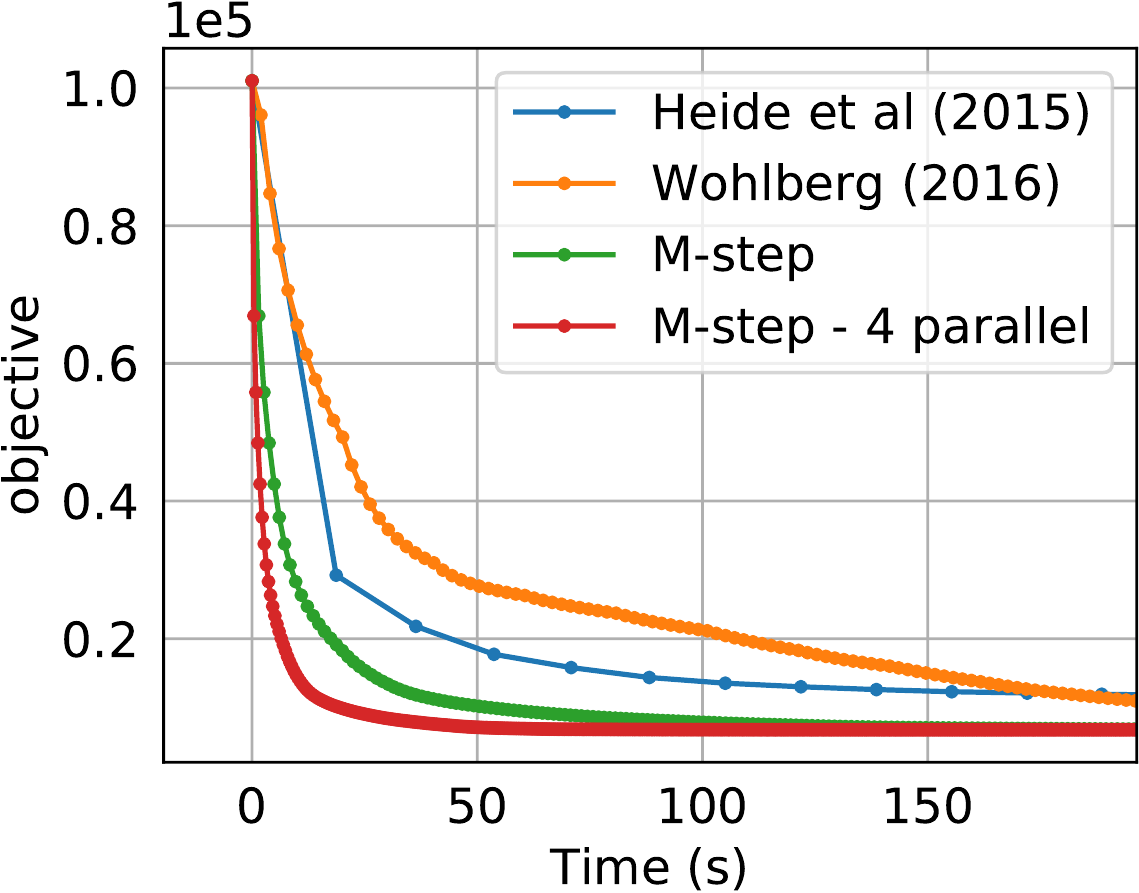}}
     \subfigure[$K=2$, $L=128$.]{
     \includegraphics[width=0.32\textwidth]{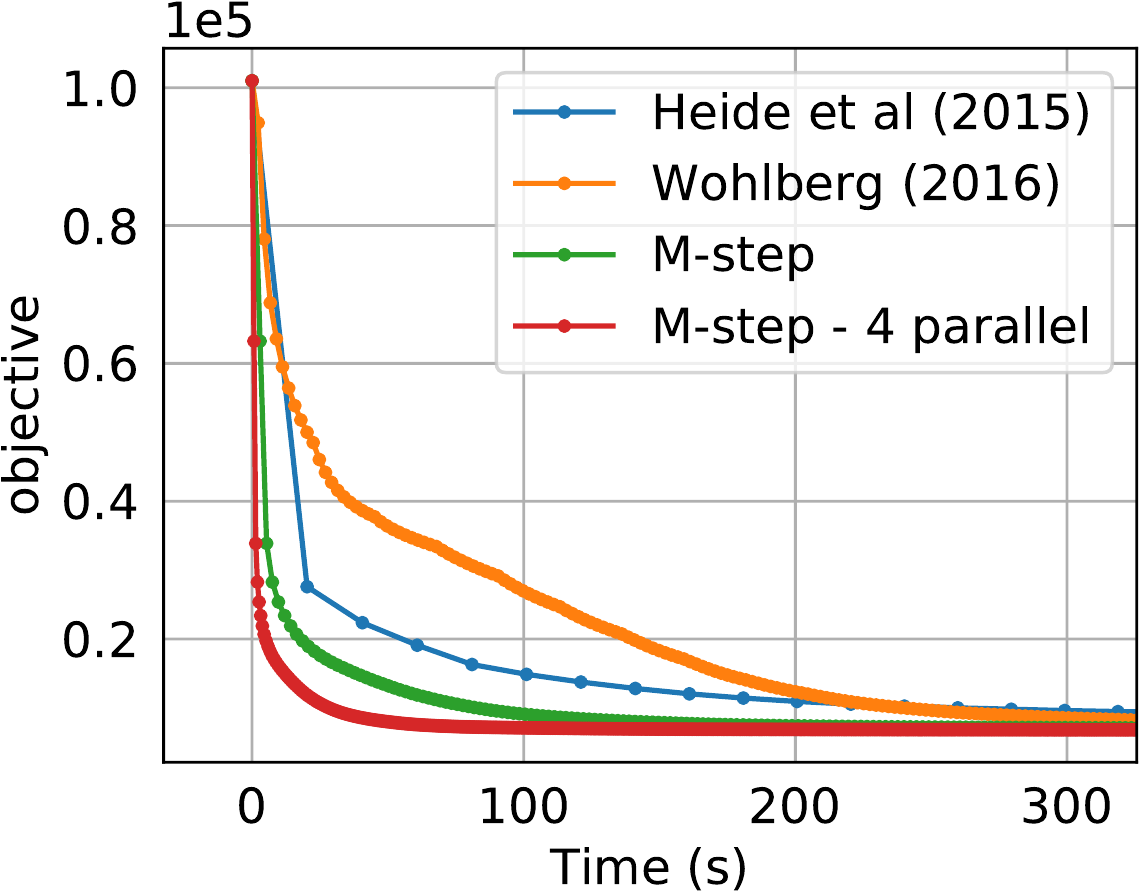}}
     \subfigure[$K=10$, $L=32$.]{
     \includegraphics[width=0.32\textwidth]{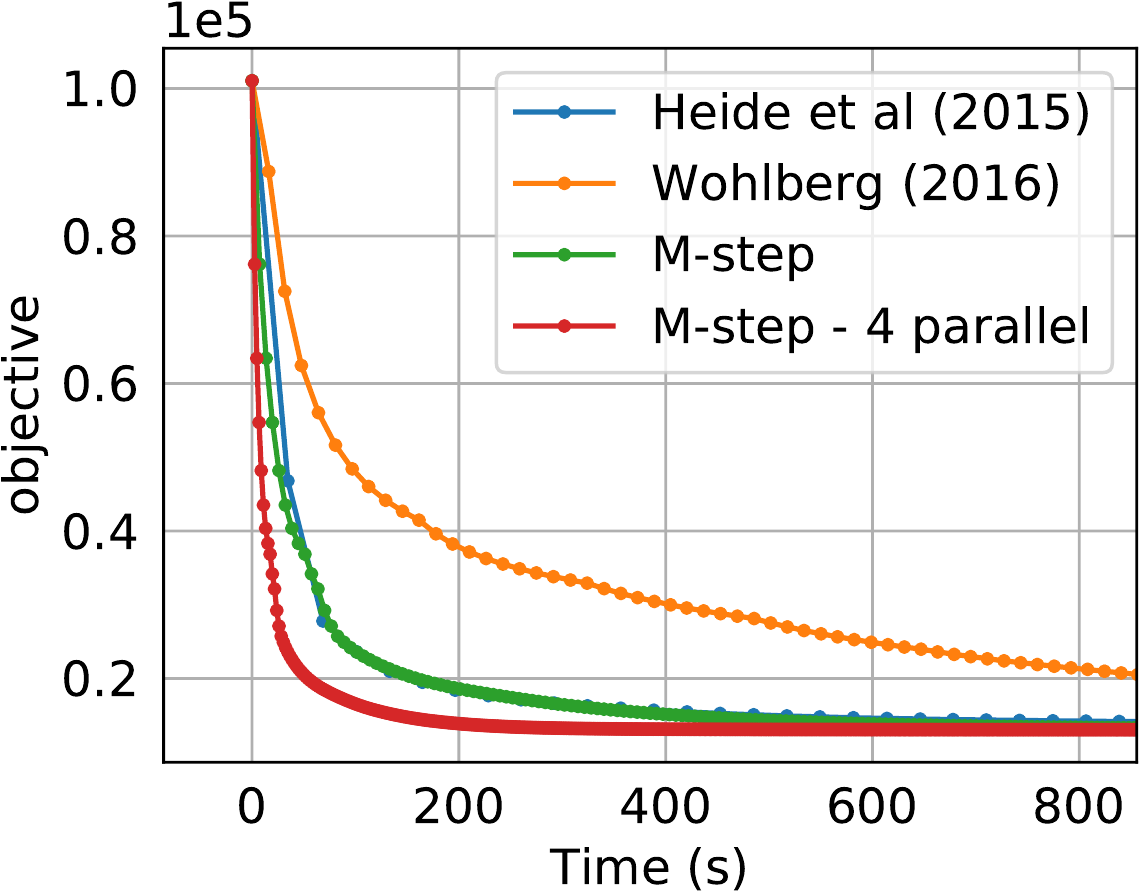}}
    \vspace{-5pt}
    \caption{Convergence of the objective function as a function of time. The y-axis shows the absolute objective function $f(x)$. Each curve is the mean over 24 different random initializations.}
    \label{fig:convergence_traditional}
\end{figure}

Next, in Fig.~\ref{fig:convergence_traditional}, we show more traditional convergence plots. In contrast to Fig.~\ref{fig:convergence} or \ref{fig:convergence_setups} where the relative objective function is shown, here we plot the absolute value of the objective function. We can now verify that each of the methods have indeed converged to their respective local minimum. Of course, owing to the non-convex nature of the problem, they do not necessarily converge to the same local minimum.

\subsection{Comparison of solver for the activations subproblem}

\begin{figure}[h]
    \centering
     \subfigure[$K=2$, $L=32$.]{
     \includegraphics[width=0.32\linewidth]{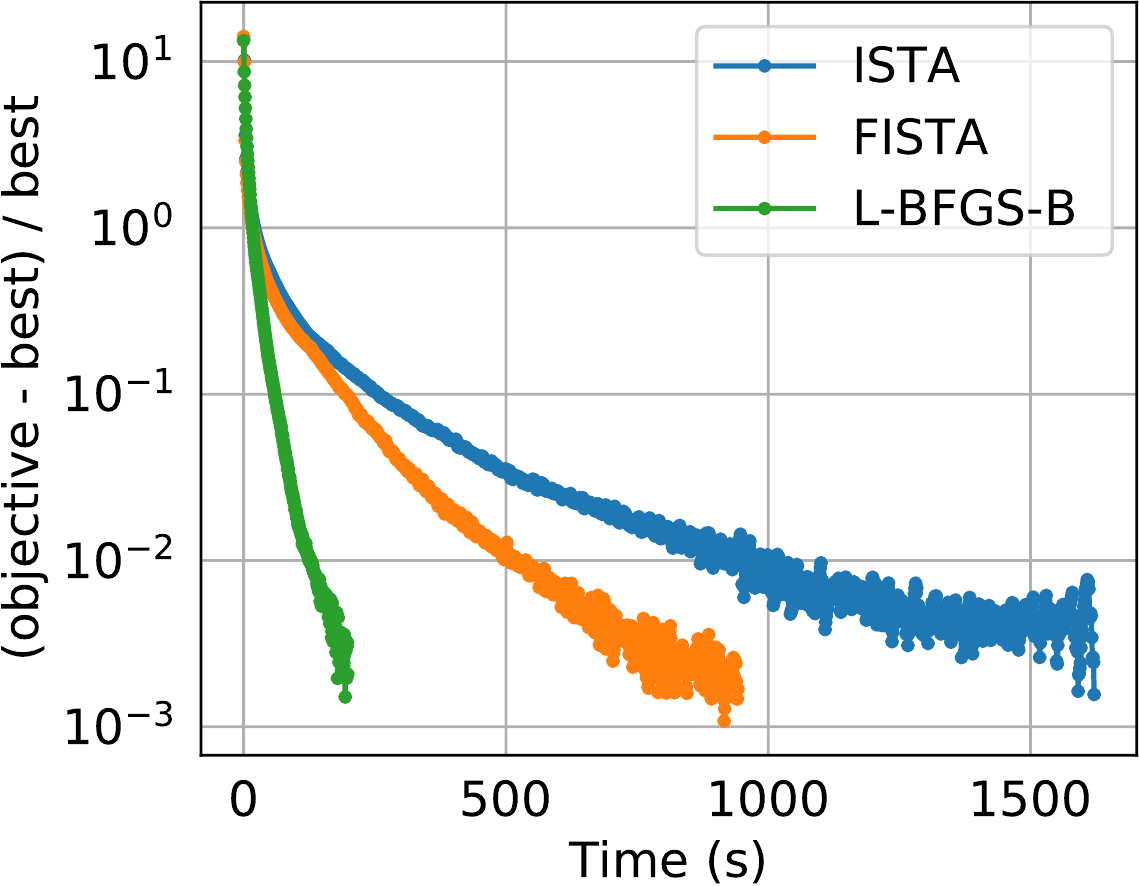}}
     \subfigure[$K=2$, $L=128$.]{
     \includegraphics[width=0.32\textwidth]{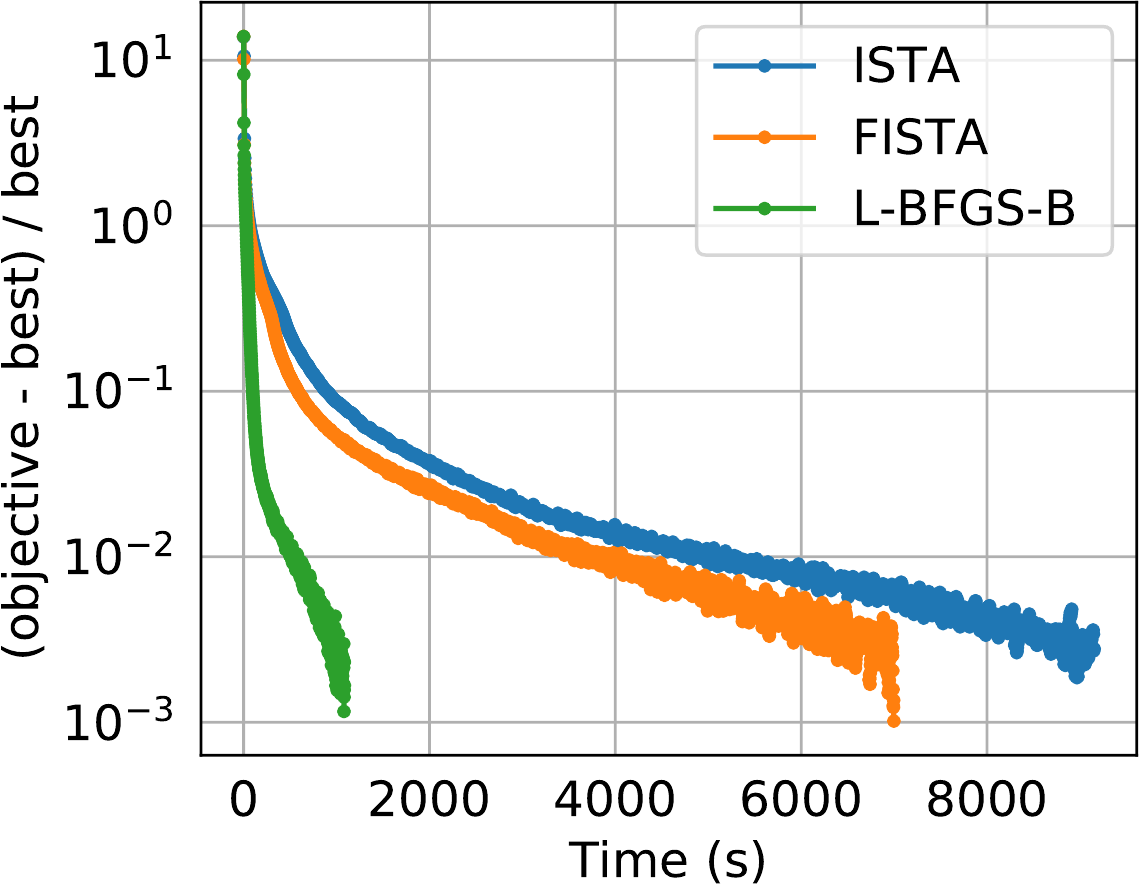}}
     \subfigure[$K=10$, $L=32$.]{
     \includegraphics[width=0.32\textwidth]{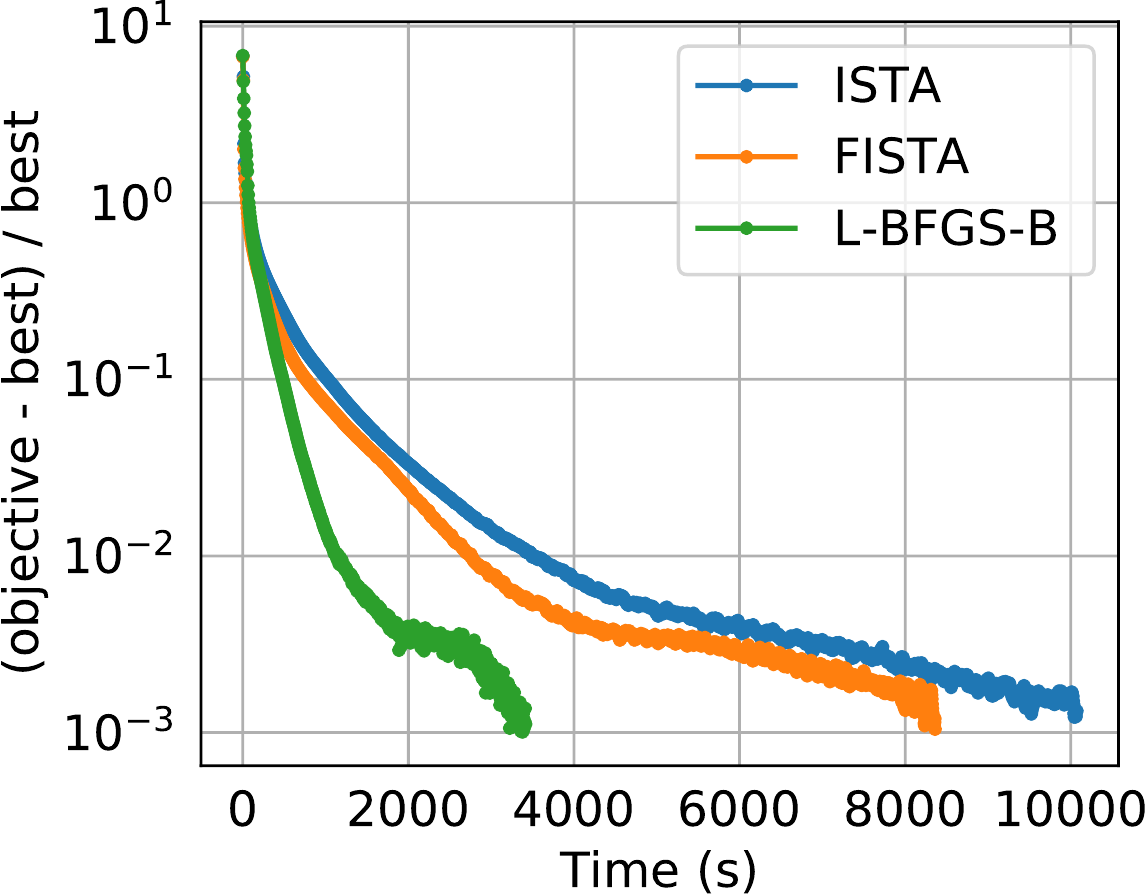}}
    \vspace{-5pt}
    \caption{Convergence speed of the relative objective function. The y-axis shows the objective function relative to the obtained minimum for each run: $(f(x) - f(x^*))/f(x^*)$. Each curve is the geometrical mean over 24 different random initializations.}
    \label{fig:convergence_z_update}
\end{figure}

Finally, we compare convergence plots of our algorithm using different solvers for the $z$-update: ISTA, FISTA, and L-BFGS-B. The rationale for choosing a quasi-Newton solver for the $z$-update becomes clear in  Fig.~\ref{fig:convergence_z_update} as the L-BFGS-B solver turns out to be computationally advantageous on a variety of setups.

\bibliographystylelatex{unsrtnat}
\newpage
\bibliographylatex{refs}

\end{document}